\documentclass[journal]{IEEEtai}

\usepackage[colorlinks,urlcolor=blue,linkcolor=blue,citecolor=blue]{hyperref}
\usepackage{amsmath,amsfonts}
\usepackage{algorithmic}
\usepackage{algorithm}
\usepackage{color,array}
\usepackage[caption=false,font=normalsize,labelfont=sf,textfont=sf]{subfig}
\usepackage{textcomp}
\usepackage{stfloats}
\usepackage{url}
\usepackage{verbatim}
\usepackage{graphicx}
\usepackage{xcolor}
\usepackage{comment}
\usepackage{multirow}
\usepackage{amsthm}
\usepackage{mathtools}
\usepackage{subfig}
\usepackage{float}

\DeclareMathOperator*{\concat}{%
    \mathchoice%
        {\Big\Vert}%
        {\big\Vert}%
        {\Vert}%
        {\Vert}%
}
\usepackage{multirow,booktabs}
\usepackage{xspace}
\newcommand{\pname}{{LFH}\xspace}
\newtheorem{myDef}{Definition}
\newtheorem{myProb}{Problem}

\newcommand*{\revise}[1]{\textcolor{black}{#1}}
\newcommand*{\revises}[1]{\textcolor{black}{#1}}
\newcommand*{\rev}[1]{\textcolor{black}{#1}}

\begin{document}

\title{Learning from Heterogeneity: A Dynamic Learning Framework for Hypergraphs}

\author{Tiehua Zhang,~\IEEEmembership{Member,~IEEE,}
        Yuze Liu,
        Zhishu Shen,~\IEEEmembership{Member,~IEEE,}
        Xingjun Ma,~\IEEEmembership{Member,~IEEE,}
        Peng Qi,~\IEEEmembership{Member,~IEEE,}
        Zhijun Ding,~\IEEEmembership{Senior Member,~IEEE,}
	Jiong Jin,~\IEEEmembership{Senior Member,~IEEE}
    
\thanks{Tiehua Zhang, Zhijun Ding are with the School of Computer Science and Technology, Tongji University, Shanghai, China (e-mail:\{tiehuaz, dingzj\}@tongji.edu.cn).}
\thanks{Yuze Liu is with Ant Group, Shanghai, China (e-mail: liuyuze.liuyuze@antgroup.com)}
\thanks{Zhishu Shen is with the School of Computer Science and Artificial Intelligence, Wuhan University of Technology, Wuhan, China (e-mail: z\_shen@ieee.org).}
\thanks{Xingjun Ma is with the School of Computer Science, Fudan University, Shanghai, China (e-mail:xingjunma@fudan.edu.cn).}
\thanks{Peng Qi is with the Department of Control Science and Engineering, Tongji University, Shanghai, China (e-mail:pqi@tongji.edu.cn).}
\thanks{Jiong Jin is with the School of Science, Computing and Engineering Technologies, Swinburne University of Technology, Melbourne, Australia (e-mail: jiongjin@swin.edu.au).}
\thanks{Corresponding author: Yuze Liu (liuyuze.liuyuze@antgroup.com)}}

\markboth{Journal of IEEE Transactions on Artificial Intelligence, Vol. 00, No. 0, Month 2020}
{T. Zhang \MakeLowercase{\textit{et al.}}: IEEE Journals of IEEE Transactions on Artificial Intelligence}

\maketitle

\begin{abstract}
Graph neural network (GNN) has gained increasing popularity in recent years owing to its capability and flexibility in modeling complex graph structure data. Among all graph learning methods, hypergraph learning is a technique for exploring the implicit higher-order correlations when training the embedding space of the graph. In this paper, we propose a hypergraph learning framework named \emph{Learning from Heterogeneity (\pname)} that is capable of dynamic hyperedge construction and attentive embedding update utilizing the heterogeneity attributes of the graph. Specifically, in our framework, the high-quality features are first generated by the pairwise fusion strategy that utilizes explicit graph structure information when generating initial node embedding. Afterwards, a hypergraph is constructed through the dynamic grouping of implicit hyperedges, followed by the type-specific hypergraph learning process. To evaluate the effectiveness of our proposed framework, we conduct comprehensive experiments on several popular datasets with twelve state-of-the-art models on both node classification and link prediction tasks, which fall into categories of homogeneous pairwise graph learning, heterogeneous pairwise graph learning, and hypergraph learning. The experiment results demonstrate a significant performance gain (an average of 12.9\% in node classification and 12.8\% in link prediction) compared with recent state-of-the-art methods.
\end{abstract}

\begin{IEEEImpStatement}
This research proposes a novel hypergraph learning framework, namely Learning from Heterogeneity (LFH), for dynamically constructing the hypergraph while adaptively learning heterogeneous hypergraph representations. LFH generates the high-quality initial node embedding using the designated pairwise fusion function, exploiting pairwise graph information at most. To encode high-order data relations, this research designs a dynamic learning process to generate different types of implicit hyperedges to construct the hyper
graph and dynamically adapts the hypergraph construction during the learning process. The framework iteratively updates node embeddings through a type-specific multi-head attention mechanism that adaptively learns the importance of heterogeneous hyperedges during the updates, thereby encoding the heterogeneity of the topology into the embedding space. LFH excels in both node classification and edge prediction tasks compared to state-of-the-art baselines.
\end{IEEEImpStatement}


\section{Introduction}~\label{sec:introduction}
\IEEEPARstart{G}{raph} learning has attracted tremendous attention in recent years owing to its prominent capability when modelling structure-based data. In particular, there has been witnessed an increasing use of graph models (i.e., GNN: Graph Neural Network) in many fields of applications, such as social network recommendation~\cite{10460339}, medical diagnosis~\cite{10510889,FedRel}, and text analysis~\cite{YuanPAMI23}. The goal of graph learning is to encode the graph structure of different input data into an embedding space, where the representations can be used for downstream node/edge/graph tasks. 

\begin{figure}[tb!]
    \centering
    \captionsetup{labelfont={color=black}}
  \includegraphics[width = 1\linewidth]{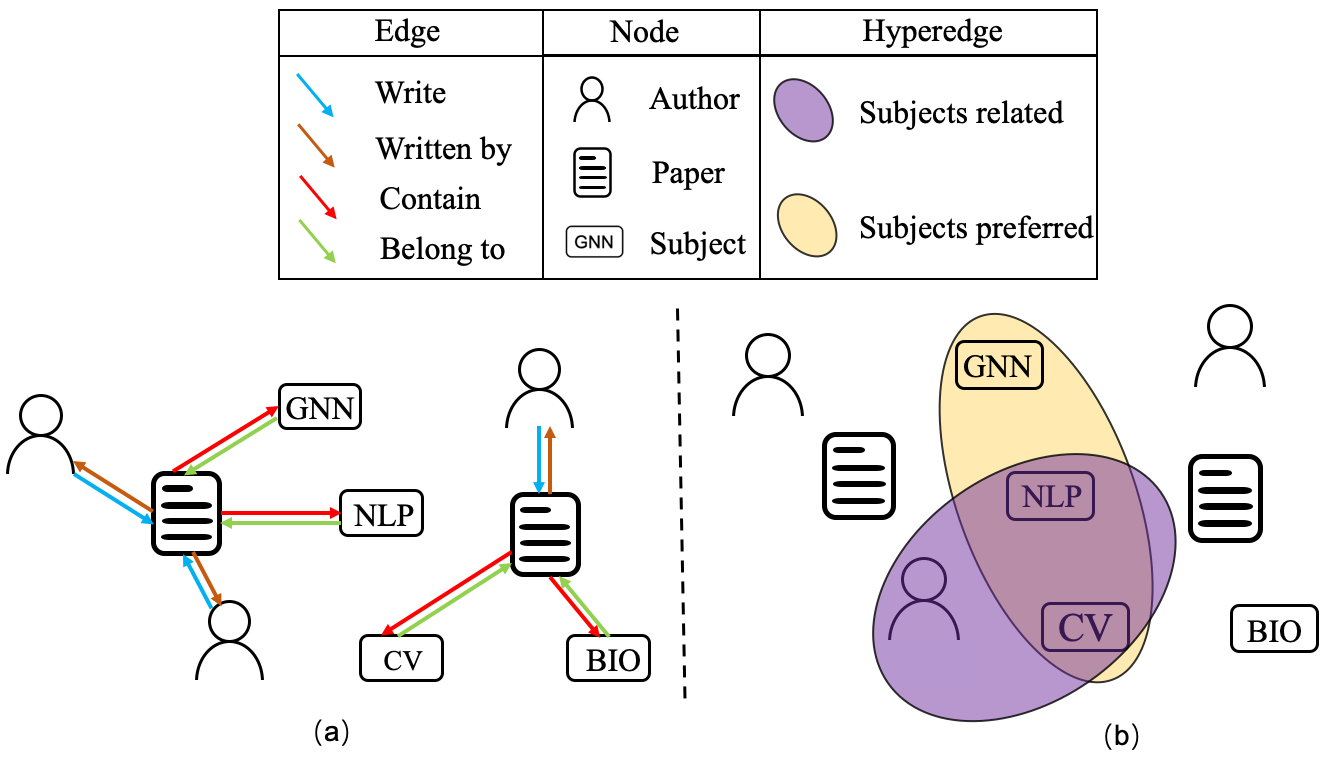}
  \caption{An illustrative example of the ACM dataset: (a) Heterogeneous pairwise graph including three node types (\emph{Author}, \emph{Paper}, \emph{Subject}) and four edge types (\emph{Write}, \emph{Written by}, \emph{Belong to}, \emph{Contain}). (b) A heterogeneous hypergraph that models two implicit data relations: \emph{preference for subjects} and \emph{the relation among the subjects}.}
  \label{fig:intro}
  \vspace{-3mm}
\end{figure}

Despite the exciting progress made by traditional GNNs, such as GCN~\cite{GCN}, GAT~\cite{GAT}, and GraphSage~\cite{hamilton2018inductive}, in learning a low-dimensional latent representation of pairwise graphs, effectively dealing with heterogeneity and capturing high-order data correlations remain critical challenges for graph learning. Taking a bibliographic network as an example,  Fig.~\ref{fig:intro} (a) sketches the ACM dataset composed of three node types (\emph{Author}, \emph{Paper}, \emph{Subject}) and four edge types (\emph{Write}, \emph{Written by}, \emph{Belong to}, \emph{Contain}). The edges in the heterogeneous pairwise graph only depict disparate relations between two nodes. These relations are often explicit and pre-defined within the dataset. For instance, the \emph{Written} relation reveals the paper's authorship, and the \emph{Belong to} relation indicates the paper's research field. The heterogeneous pairwise graph can be modeled by most of the aforementioned traditional GNNs and heterogeneous graph learning methods, such as PC-HGN~\cite{PC-HGN}, HAN~\cite{WangWWW19} and HGT~\cite{HuWWW20}. However, such pairwise relations may fail to uncover underlying high-order relations, such as which subjects are more relevant to an author. Pairwise graph learning methods, in this case, become intractable. By contrast, hypergraph presents more complex non-pairwise relationships, where these implicit relationships are not only dyadic (pairwise) but rather triadic or even higher~\cite{Berge1984}. Specifically, a heterogeneous hypergraph is capable of modeling a more complex graph structure by constructing different types of hyperedges, in which every hyperedge can connect any number of nodes to model a specific type of data relationship. Thus, the hyperedges in the heterogeneous hypergraph are able to capture the implicit relations by clustering different groups of nodes, from which high-order information can be better encoded. As shown in Fig.~\ref{fig:intro} (b), the purple hyperedge models an author's preference for subjects, while the yellow one models the relation among different subjects. In general, hypergraph learning can be regarded as a generalized form of traditional graph learning.

Owing to the great potential of exploring high-order correlations among the data, hypergraph learning has drawn increasing attention from the academic community and has been employed in applications such as image classification~\cite{wang2015visual}, video segmentation~\cite{Yan_2020_CVPR}, and hyperspectral image analysis~\cite{LuoTGRS20}. 
How to construct hyperedges lies at the core of hypergraph learning. Solutions to hypergraph construction in the existing works are generally divided into two categories, i.e., attribute-based construction and neighborhood-based construction. The early work focused on constructing the hypergraph based on the features of each vertex \cite{gao2011tag}. Recently, heterogeneous-hypergraph-based models (HeteHG-VAE~\cite{FanPAMI22}, Meta-HGT~\cite{liu2023meta}) utilize pre-defined relations, such as multiple authors co-authoring a paper in citation dataset, to construct the heterogeneous hypergraph. This strategy instructs the model to use only the feature information which inevitably sacrifices certain generality from the topological perspective, and makes the model inapplicable in other datasets. The neighborhood-based strategy addresses this weakness by enabling the locality concept. Specifically, a hyperedge is constructed through clustering a master vertex and its neighbor vertices~\cite{GaoTPAMI22}. However, the clustering method (e.g., \textit{k}-nearest neighbors (\textit{k}-NN)~\cite{FengAAAI19}) separates hypergraph construction from the graph learning process, and thus is sensitive to noisy data, especially in visual classification tasks. Moreover, most neighborhood-based models fail to incorporate topological heterogeneity into the learning process, which limits the learning to a single type of hyperedges. In order to overcome this, general hypergraph neural network (HGNN+)~\cite{GaoTPAMI23} designs four types of hyperedge generation approaches, including pairwise edge-based, attribute-based, \textit{k}-hop neighbors-based and feature-based, to form the hypergraph. Nevertheless, HGNN+ separates the hypergraph construction from the graph learning process and only uses a trainable vector to represent the weight of all hyperedges (i.e., a weighting scalar for each hyperedge), without considering the heterogeneity information. Heterogeneity modelling is another challenge in hypergraph learning~\cite{antelmi2023survey}. Existing heterogeneous hypergraph models like HHGSA~\cite{khan2023heterogeneous} and HWNN~\cite{sun2021heterogeneous} treat the heterogeneous hypergraph as multiple homogeneous hypergraphs, without leveraging the cross-type interactions among diverse hyperedge and node types.
Additionally, hypergraph learning (HL)~\cite{HL} models the representation of hypergraph by formulating an optimization problem, which partitions nodes into several parts and constructs the objective function to minimize the supervised loss and the partition cost when generating hyperedges. Yet, it is proven that the optimization is NP-complete, leading to a high computational complexity~\cite{FengAAAI19}. 

To tackle these challenges, we propose a dynamic learning framework, \emph{Learning from Heterogeneity (\pname)}, to improve the quality of representation learning for hypergraphs. Our \pname framework is composed of three key modules: 1) initial embedding generation, 2) dynamic hypergraph construction, and 3) attention-based heterogeneous hypergraph learning. Concretely, initial embedding generation aims to exploit the pairwise connectivities in the graph, helping to fuse the explicit topological information into the initial embedding space. Following that, as an integral part of representation learning, the heterogeneous hypergraph is constructed and then learned dynamically. 

In summary, the main contributions of this work are:
{\color{black}
\begin{itemize}
    \item We propose a novel hypergraph learning framework \pname with three modules: initial embedding generation, dynamic hypergraph construction, and attention-based heterogeneous hypergraph learning. A pairwise fusion strategy is also considered to fully exploit the explicit pairwise graph information to generate high-quality initial embedding.
    
    \item We design a dynamic learning process to generate different types of implicit hyperedges to construct the hypergraph and dynamically adapt the hypergraph construction during the learning process. The embedding is updated iteratively through a type-specific multi-head attention mechanism, which adaptively learns the importance of heterogeneous hyperedges during node embedding updates, thereby encoding the heterogeneity of the topology into the embedding space.
    \item We conduct extensive experiments on two different downstream tasks, including node classification and linked prediction, and show that our \pname outperforms state-of-the-art homogeneous pairwise graph learning models, heterogeneous pairwise graph learning models, and hypergraph learning models consistently across all datasets by a large margin.
    \item We have made our source code publicly available to contribute further to advancements in this field. The source code is available at https://github.com/paperCodeEric/LFH.
\end{itemize}
}
The remainder of this paper is organized as follows. Section~\ref{sec:relatedwork} reviews the related works of graph learning on homogenous, heterogenous, and hypergraphs. Section~\ref{sec:preliminaries} provides the preliminary knowledge of hypergraphs, while Section~\ref{sec:methodology} introduces the detail of our proposed \pname framework, along with analyzes and discussions. The experimental results are reported and analyzed in Section~\ref{sec:experiment}. Finally, Section~\ref{sec:conclusion} concludes the paper.

\section{Related Work}~\label{sec:relatedwork}
\vspace{-6mm}
\subsection{Homogeneous Graph Learning}
As a basic graph structure, a homogeneous graph consists of a single node type with a single relation. Typical GNN models for homogeneous graphs include GCN~\cite{GCN} and GAT~\cite{GAT}. GCN extends the traditional convolutional neural networks (CNN) to handle graph-structured data by performing convolutions in the spectral/spatial domain of the graph nodes to capture the structural information of the graph. On the other hand, GAT applies the self-attention mechanism to assign dynamic weights for the neighbors of a node and then take a weighted sum of their embeddings to obtain the node's representation. This enables GAT to learn adaptive neighborhood representations and capture complex relationships between nodes.

The primary difference between various GNN models lies in the way how messages are passed between the nodes to learn the representation.
Instead of training individual embeddings for each node, GraphSAGE~\cite{hamilton2018inductive} leverages the node features in the learning algorithm to train a set of aggregator functions to generate embeddings for entirely unseen nodes. Aiming for visual question answering (VQA) services, Li \textit{et al.} proposed a GAT-based platform that encodes each image into a graph and models multi-type inter-object relations to learn relation representation from the graphs~\cite{LiICCV2019}. Jiang \textit{et al.} presented a GCN-based framework (GLCN) for graph data representation learning. GLCN generates similarity-based graph structure by simultaneous graph learning and graph convolution in a unified network architecture~\cite{JiangCVPR19}. Compared with the embedding frameworks that can only generate embeddings for a single fixed graph like transductive learning in GCN, Zeng \textit{et al.} proposed GraphSAINT, a graph sampling based inductive representation learning method, to generalize across different graphs.  GraphSAINT generates low-dimensional vector representations for the nodes and is beneficial for large graphs with rich node attribute information~\cite{zeng2020graphsaint}. 


\subsection{Heterogeneous Graph Learning}
Different from homogeneous graphs, the nodes in a heterogeneous graph are usually connected with various types of neighbors via different types of relations. As such, representation learning on heterogeneous graphs is much more challenging. It needs to not only incorporate heterogeneous structure (graph) information but also consider the heterogeneous attributes associated with each node~\cite{WangTBD22}.

Zhang \textit{et al.} proposed a heterogeneous GNN model (HetGNN) for capturing both structure and content heterogeneity. HetGNN first captures the strongly correlated heterogeneous neighbors of each node and then aggregates feature information of these sampled neighbors~\cite{ZhangKDD19}. Zhao \textit{et al.} proposed a GNN-based framework named HGSL that jointly performs heterogeneous graph structure learning and GNN parameter learning for classification. In HGSL, the feature similarity graphs, the feature propagation graphs, and the semantic graphs are generated separately so as to comprehensively learn an optimal heterogeneous graph~\cite{ZhaoAAAI21}. Zhang \textit{et al.} designed a relation-centered pooling and convolution (PC-HGN) operation that enables relation-specific sampling and cross-relation convolutions on heterogeneous graphs, from which the structural heterogeneity of the graph can be better encoded into the embedding space through the adaptive training process~\cite{PC-HGN}. HAN (Heterogeneous graph Attention Network)~\cite{WangWWW19} is a heterogeneous graph learning framework based on node-level and semantic-level attention mechanisms. Specifically, node-level attention learns the importance between a node and its meta-path based neighbors, while semantic-level attention learns the importance of different meta-paths. Hu \textit{et al.} presented a heterogeneous graph transformer (HGT) for modeling heterogeneous graphs~\cite{HuWWW20}. HGT introduces the node-type and edge-type dependent attention mechanism while different trainable parameters are assigned to each node and edge type. HGT can incorporate high-order heterogeneous neighbor information, which automatically learns the importance of implicit meta path. Considering relation-aware characteristics, Yu \textit{et al.} proposed a relation-aware representation learning model for heterogeneous graphs (R-HGNN). This model derives a fine-grained representation from a group of relation-specific node representations reflecting the characteristics of the node associated with a specified relation~\cite{YuTKDE22}. 



\subsection{Hypergraph Learning}

Hypergraph learning explores the high-order correlations in the data, which extends the traditional graph learning models to a high dimensional and nonlinear space. For example, Sun \textit{et al.} designed a hypergraph framework for human behavioral analysis to indicate the interactions between individuals and their environments. Since each edge in the hyperedges contains multiple nodes information, hypergraph learning is suitable to analyze rich sociological criteria effectively~\cite{SunTKDE23}. As a result, hypergraph learning offers a promising solution for analyzing complex structured data with satisfactory performance in practice~\cite{GaoTPAMI22}. The construction of the hypergraph from the given data is a key step for hypergraph learning, which significantly affects the final learning performance.

Guo \textit{et al.}  designed a representation learning framework to enhance the learning of individuals’ preferences from their friends’ preferences for recommendation system. To exploit the group similarity for an accurate group representation, all groups are connected as hypergraph, and the task of group preference learning are treated as embedding hyperedges in a hypergraph, where an inductive hyperedge embedding method is introduced~\cite{GuoTIS21}. Zhang \textit{et al.} proposed dynamic hypergraph structure learning (DHSL) to update hypergraph structure iteratively during the learning process~\cite{DHSL}. It is essential to make dynamic modifications to the initial hypergraph structures from adjusted feature embedding. Feng \textit{et al.} presented a framework named hypergraph neural network (HGNN) for handling complex and high-order correlations. In HGNN, the complex data correlation is formulated in a hypergraph structure, and a hyperedge convolution operation is used to exploit the high-order data correlation for representation learning~\cite{FengAAAI19}.

To exploit high-order relations among the features, Jiang \textit{et al.} proposed a dynamic hypergraph neural networks (DHGNN) framework that is composed of dynamic hypergraph construction (DHG) and hypergraph convolution (HGC). DHG utilizes the \textit{k}-NN method to generate the basic hyperedge and extends the adjacent hyperedge set by \textit{k}-means clustering, with which the local and global relations can be extracted. HGC is designed to encode high-order data relations in the hypergraph structure~\cite{Jiangijcai2019}. Luo \textit{et al.} proposed a hypergraph framework named SHINE, which can effectively learn subgraph representations by simultaneously utilizing the learned representations and inductively infer representations for subgraph predictions for genetic medicine~\cite{LuoNOPS22}. Cai \textit{et al.} introduced a hypergraph structure learning (HSL) framework to optimize the hypergraph structure and HGNNs simultaneously in an end-to-end way. To efficiently learn the hypergraph structure, HSL adopts a hyperedge sampling strategy to prune the redundant hyperedges, which is followed by an incident node sampling for pruning irrelevant incident nodes and discovering potential implicit connections. The consistency between the optimized structure and the original structure is maintained by the intra-hyperedge contrastive learning module~\cite{Caiijcai2022}. Gao \textit{et al.} proposed a tensor-based dynamic hypergraph learning (t-DHL) model to efficiently learn dynamic hypergraphs. t-DHL utilizes a tensor representation to characterize the dynamic hypergraph structure more flexibly. During the optimization of the tensor representation, not only the weights but also the number and order of hyperedges can be adjusted. As an extended version of HGNN, a general hypergraph neural network framework named HGNN+ was proposed for modeling high-order representation among the data, which is achieved by bridging multi-modal/multi-type data and hyperedge with hyperedge groups~\cite{GaoTPAMI23}. Liu \textit{et al.} proposed a meta-path-aware hypergraph transformer (Meta-HGT) to encode heterogeneous information network embedding. Meta-HGT constructs the heterogeneous hypergraph based on pre-defined meta-paths, such as multiple authors co-authoring a paper, in the citation graph~\cite{liu2023meta}. Fan \textit{et al.} introduced a heterogeneous hypergraph variational autoencoder (HeteHG-VAE) to learn the latent topology information of the heterogeneous hypergraph, which is constructed from pre-defined high-order relations, specifically for the link prediction task. HeteHG-VAE designs a hyperedge attention module that employs a non-linear attention scheme. This scheme learns the importance for different types of nodes within each hyperedge during the process of encoding the hyperedge embedding~\cite{FanPAMI22}. Khan \textit{et al.} proposed a heterogeneous hypergraph neural network for social recommendation using attention network (HHGSA), which treats the heterogeneous hypergraph as multiple homogeneous hypergraphs. HHGSA directly employs the hypergraph attention network (HGAT)~\cite{hyperatt} to encode node and hyperedge embeddings, and then obtains the final representations of users and items through simple aggregation~\cite{khan2023heterogeneous}.

\rev{To elucidate the tasks addressed in relevant works, hypergraph learning methods can be divided into two categories: methods for relational data and non-relational data. The methods for relational data focus on tasks that naturally involve explicit correlations among data, with well-known examples including social network tasks~\cite{Caiijcai2022,liu2023meta,FanPAMI22} and social network tasks~\cite{SunTKDE23,GuoTIS21,Jiangijcai2019}. Previous works usually define handcrafted rules~\cite{GuoTIS21,liu2023meta,FanPAMI22} or utilize a similarity function to select neighboring nodes based on existing pairwise relations to construct the hypergraph~\cite{SunTKDE23,Jiangijcai2019,Caiijcai2022}. On the other hand, methods~\cite{DHSL,FengAAAI19,hyperatt} for non-relational data focus on datasets containing independent data points, such as images, where there are no explicit correlations among different images. These methods utilize unsupervised clustering strategies to construct hypergraphs, from which hypergraph learning models can be applied.}

Recent works mostly focus on neighborhood-based construction strategies when constructing hyperedges in the hypergraph. Specifically, hyperedge is often constructed through clustering a master vertex and its neighbor vertices~\cite{Zhang2020HypergraphLP}. However, for the commonly adopted clustering methods like \textit{k}-NN~\cite{wang2015visual,Jiangijcai2019,FengAAAI19}, they treat hypergraph construction and graph learning separately. As a result, these models are susceptible to noisy data which limits their applications. Apart from that, most neighborhood-based models fail to incorporate the heterogeneity of the topology, leading to a single type hyperedge~\cite{GaoTPAMI22}. In this work, we propose a heterogeneous hypergraph learning framework \pname to solve the above mentioned issues. \pname constructs the hyperedge based on different edge types, which can then be integrated into the learning process to improve the learned representations.


\section{Preliminaries and problem formulation}
~\label{sec:preliminaries}
In this section, we first introduce the relevant definitions and then formulate the key problems associated with representation learning on heterogeneous hypergraphs. The notations used in this paper are listed in Table~\ref{tab:notation}.

\begin{myDef}
\textbf{(Heterogeneous Pairwise Graph)}. A heterogeneous graph $\dot{\mathcal{G}} =\{ \dot{\mathcal{V}},\dot{\mathcal{E}},\dot{\mathcal{T}_{e}},\dot{\mathcal{T}_{v}}\}$ is defined as a graph with multiple node types $\dot{\mathcal{T}_{v}}$ or edge types $\dot{\mathcal{T}_{e}}$, where $\dot{\mathcal{V}} = \{\emph{v}_{1},\emph{v}_{2},...,\emph{v}_{N}\}$ denotes the set of nodes and $\dot{\mathcal{E}} = \left\{\left(\emph{v}_{i},\emph{v}_{j}\right)|\emph{v}_{i},\emph{v}_{j}\in\dot{\mathcal{V}}\right\}$ denotes the set of edges. We define the node type mapping function $\dot{\phi}:\dot{\mathcal{V}}\rightarrow\dot{\mathcal{T}_{v}}$ and the edge type mapping function $\dot{\psi}:\dot{\mathcal{E}}\rightarrow\dot{\mathcal{T}_{e}}$, respectively.
\end{myDef}
\begin{myDef}
\textbf{(Heterogeneous Hypergraph)}. A heterogeneous hypergraph is defined as $\mathcal{G} =\{\mathcal{V},\mathcal{E},\mathcal{T}_{e},\mathcal{T}_{v},\mathcal{W}\}$, where $\mathcal{V} = \{\emph{v}_{1},...,\emph{v}_{N}\}$ is the set of nodes and $\mathcal{E} = \{\emph{e}_{1},...,\emph{e}_{M}\}$ is the set of hyperedges. $\emph{N}$ and $\emph{M}$ are the number of nodes and hyperedges, respectively. For any hyperedge $\emph{e} \in \mathcal{E}$, it is composed of a subset of nodes $\mathcal{V}$, which can be defined as $\emph{e} = \{\emph{v}_{i},...,\emph{v}_{j}\}\subseteq\mathcal{V}$. We define the node type mapping function $\phi:\mathcal{V}\rightarrow\mathcal{T}_{v}$ and the hyperedge type mapping function $\psi:\mathcal{E}\rightarrow\mathcal{T}_{e}$, respectively. Each heterogeneous hypergraph presents multiple hyperedge types, i.e., $|\mathcal{T}_{e}| > 1$. A positive diagonal matrix $\mathcal{W}\in\mathcal{R}^{M\times M}$ is used to denote the weight of each hyperedge, that is, $diag\left(\mathcal{W}\right) = \left[w\left(\emph{e}_{1}\right),...,w\left(\emph{e}_{M}\right)\right]$. The relationship between nodes and hyperedges is usually represented by incidence matrix $\emph{H}\in\mathcal{R}^{N\times M}$, with entries
are continuous numbers within the range of $[0,1]$. If the hyperedge $e_{j}$ is incidenced with node $v_{i}$, the value of $\emph{H}\left(v_{i},e_{j}\right)$ is positive ; otherwise, $\emph{H}\left(\emph{v}_{i},\emph{e}_{j}\right) = 0$. The element $\emph{H}\left(\emph{v}_{i},\emph{e}_{j}\right)$ can be regarded as the importance of node $\emph{v}_{i}$ for hyperedge $\emph{e}_{j}$. We can define the degree of each node $\emph{v}_{i}\in\mathcal{V}$ and the edge degree of hyperedge $\emph{e}_{j}\in\mathcal{E}$ as:
\begin{equation}
    d\left(\emph{v}_{i}\right) = \sum_{\emph{e}_{p}\in\mathcal{E}}w\left(\emph{e}_{p}\right)\cdot\emph{H}\left(\emph{v}_{i},\emph{e}_{p}\right)
\end{equation}
\begin{equation}
    \delta\left(\emph{e}_{j}\right) = \sum_{\emph{v}_{p}\in\mathcal{V}}\emph{H}\left(\emph{v}_{p},\emph{e}_{j}\right)
\end{equation}
\end{myDef}


\begin{figure*}[tb!]
    \vspace{-2mm}
    \captionsetup{labelfont={color=black}}
    \centering
    \includegraphics[width=0.8\linewidth]{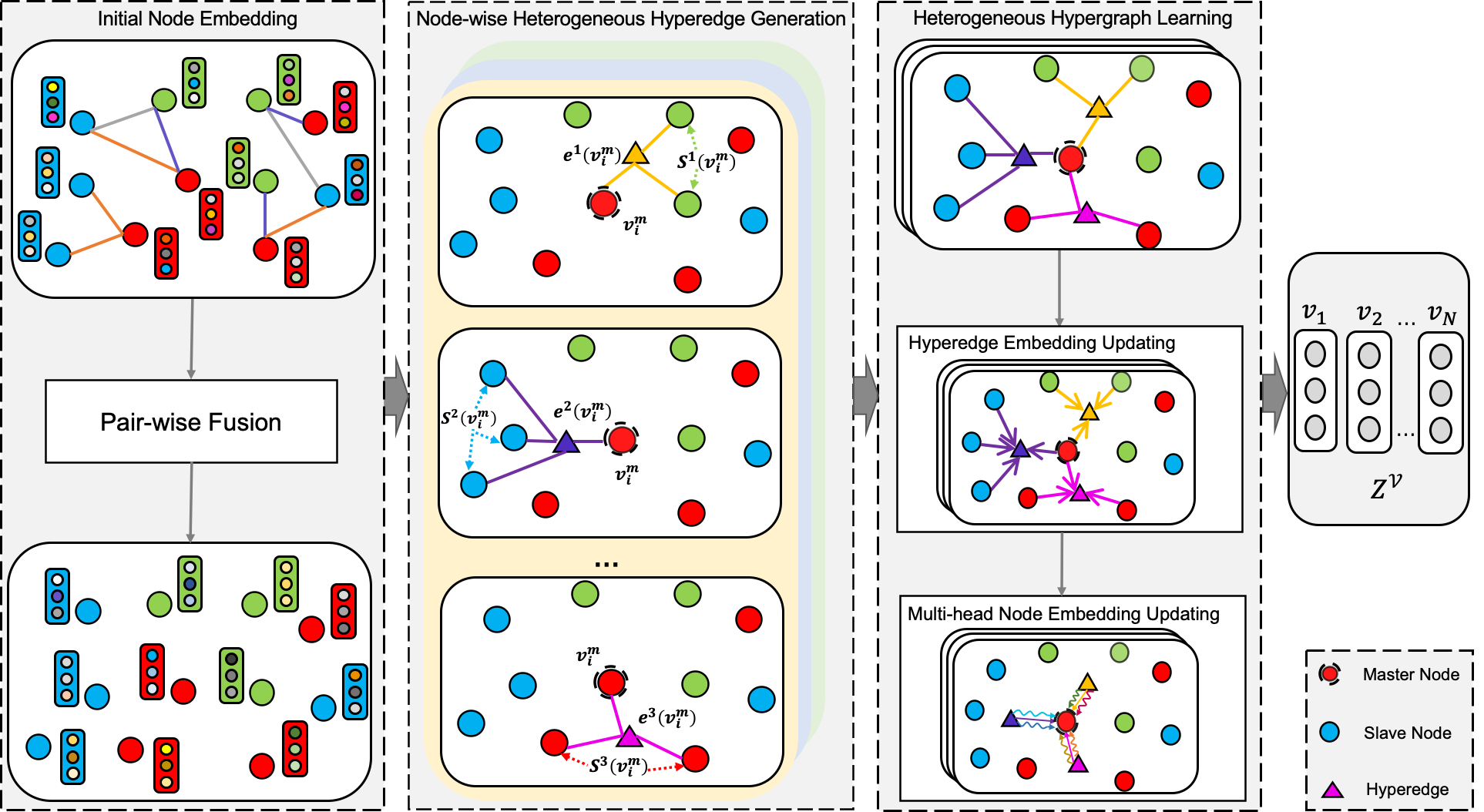}
  \caption{An overview of our proposed \pname framework.}
  \vspace{-5mm}
  \label{fig:overview} 
\end{figure*} 
\vspace{-2mm}

\begin{myProb}
\textbf{(Hyperedge Generation in Heterogeneous Hypergraph)}.
While node set $\mathcal{V}$ explicitly exists in graph data, the hyperedge set $\mathcal{E}$ is normally implicit. Following Definition 2, any hyperedge $\emph{e}\in\mathcal{E}$ contains a subset of nodes $\mathcal{V}$. We aim to learn a hyperedge construction function $f_{con}:\mathcal{V}\rightarrow\mathcal{E}$, where hyperedges are constructed based on the given nodes. Moreover, this function can be used for constructing incidence matrix \emph{H}.

\end{myProb}

\begin{myProb}
\textbf{(Representation Learning on Heterogeneous Hypergraph)}.
Given a heterogeneous hypergraph  $\mathcal{G} =\{\mathcal{V},\mathcal{E},\mathcal{T}_{e},\mathcal{T}_{v},\mathcal{W}\}$, we aim to learn a mapping function $f_{emb}\left(\mathcal{G}\right)\rightarrow\emph{Z}^{\mathcal{G}}$, where $\emph{Z}^{\mathcal{V}}\in\mathcal{R}^{d\times N}$ indicates representation embedding of all nodes in $\mathcal{G}$. This representation embedding can be used for downstream predictive tasks such as node classification and link prediction. 
\end{myProb}

\begin{table}[t!]
\vspace{-2mm}
\caption{Table of key notations.}
\vspace{-1mm}
\centering
\resizebox{\linewidth}{!}{
\begin{tabular}{c||l}
\hline
Notation & \ Description\\
\hline
\hline
$\|$ & Concatenation \\
$\|\cdot\|_{1}$ & $l_{1}$ norm \\
$\|\cdot\|_{2}$ & $l_{2}$ norm \\
$|\cdot|$ & The set size \\
$\mathcal{L}_{recon}$ & Reconstruction loss\\
$\mathcal{L}_{label}$ & Label loss\\
$Softmax\left(\cdot\right)$ & Softmax function \\
$f_{con}$ & Hyperedge construction function \\
$f_{emb}$ & Representation embedding mapping function\\
$\Theta_{p}$,$\theta$,$\Theta^{att}$ & Learnable model parameters \\
$\dot{\mathcal{G}}$ & Heterogeneous pairwise graph \\
$\dot{\mathcal{V}}$ & The set of nodes in the heterogeneous pairwise graph\\
$\dot{\mathcal{E}}$ & The set of edgees in the heterogeneous pairwise graph \\
$\dot{\mathcal{T}_{v}}$ &  The set of node types in the heterogeneous pairwise graph\\
$\dot{\mathcal{T}_{e}}$ &  The set of edge types in the heterogeneous pairwise graph\\
$\dot{\emph{X}}$ & Raw node features in the heterogeneous pairwise graph \\
$\dot{\emph{x}_{i}}$ & Raw feature of node $v_{i}$ in the heterogeneous pairwise graph\\
$\dot{\phi}$ & Node type mapping function in the heterogeneous pairwise graph\\
$\dot{\psi}$ & Edge type mapping function in the heterogeneous pairwise graph\\
$\mathcal{G}$ & Heterogeneous hypergraph \\
$\mathcal{V}$ & The set of nodes in the heterogeneous hypergraph\\
$\mathcal{E}$ & The set of edgees in the heterogeneous hypergraph \\
$\mathcal{T}_{v}$ &  The set of node types in the heterogeneous hypergraph\\
$\mathcal{T}_{e}$ & The set of edge types in the heterogeneous hypergraph\\
$\phi$ & Node type mapping function in the heterogeneous hypergraph\\
$\psi$ & Edge type mapping function in the heterogeneous hypergraph\\
$\emph{X}$ & Initial node embedding in the heterogeneous hypergraph\\
$\emph{x}_{i}$ & Initial node embedding of node $v_{i}$ in the heterogeneous hypergraph \\
$\mathcal{W}$ & Weight matrix of hyperedges\\
$\emph{H}$ & Incidence matrix \\
$\emph{Z}^{\mathcal{V}}$ & Embedding of all nodes in $\mathcal{V}$ \\
$\emph{Y}^{\mathcal{V}}$ & Labels of nodes \\
$v$ & A node $v\in\mathcal{V}$ \\
$e$ & A node $e\in\mathcal{E}$ \\
$w\left(e\right)$ & The weight of hyperedge $e$\\
$d\left(v\right)$ & Degree of node $v$ \\
$\delta\left(e\right)$ & Degree of edge $e$ \\
$v_{i}^{m}$ & The master node of $v_{i}^{m}$ \\
$\emph{S}\left(v_{i}^{m}\right)$ & The slave node set of the master node $v^{m}_{i}$ \\
$\Tilde{\emph{S}}\left(v_{i}^{m}\right)$ & The candidate slave node set of the master node $v^{m}_{i}$ \\
$\mathcal{T}\left(\Tilde{\emph{S}}\left(v_{i}^{m}\right)\right)$ & The type of the candidate slave node set \\
$\hat{\emph{X}}\left(v_{i}^{m}\right)$ & The reconstructed vector of the master node $v_{i}^{m}$\\
$\emph{p}_{v_{i}^{m}}$ & The reconstruction coefficient vector of $v_{i}^{m}$ \\
$c\left(v^{m}_{i}\right)$ & Reconstruction loss of the master node $v^{m}_{i}$ \\
$M$ & The number of hyperedges\\
$N$ & The number of nodes \\
$n_{i}$ & The number of candidate slave node set associated with node $v_{i}$ \\
$M_{i}$ & The number of hyperedges associated with the node $v_{i}$ \\
$\emph{E}$ & Hyperedges embeddding\\
\rev{$H$} & The number of attention heads\\
\hline
\end{tabular}
}
\vspace{-6mm}
\label{tab:notation}
\end{table}

\section{Heterogeneous Hypergraph}~\label{sec:methodology}
In this section, we introduce our proposed \emph{Learning from Heterogeneity} (\pname) framework for hypergraphs. \figurename~\ref{fig:overview} provides an overview of \pname.
We first demonstrate the pairwise fusion process from the heterogeneous pairwise graph to obtain a high-quality initial node embedding. 
Afterward, we dynamically generate the heterogeneous hyperedges among the nodes with different node types through feature reconstruction. After the heterogeneous hypergraph is constructed, our proposed dynamic hypergraph learning can then be performed to train the node embedding space, in which node representations can be derived for downstream node classification and link prediction tasks.

\subsection{Initial Node Embedding}
As demonstrated in Fig.~\ref{fig:overview}, the input of this step is the heterogeneous pairwise graph. \rev{It is of great importance to generate high-quality initial node embedding to better exploit the implicit data relations in the hypergraph~\cite{GaoTPAMI23,FanPAMI22}. Intuitively, fusing pairwise information from a heterogeneous pairwise graph into the initial node embeddings can help capture high-order correlations among the nodes and thus affects the construction of the hypergraph. }Given a heterogeneous pairwise graph $\dot{\mathcal{G}} =\{ \dot{\mathcal{V}},\dot{\mathcal{E}},\dot{\mathcal{T}_{e}},\dot{\mathcal{T}_{v}}\}$ and its raw node features $\dot{\emph{X}} = \left[\dot{\emph{x}}_{1},\dot{\emph{x}}_{2},...,\dot{\emph{x}}_{N}\right]\in\mathcal{R}^{D\times N}$, where the raw feature of node $v_{i}\in\mathcal{V}$ is $\dot{\emph{x}}_{i}\in\mathcal{R}^D$. 
The pairwise fusion can be implemented by any GNN model that operates on heterogeneous pairwise graphs. Formally, the pairwise fusion process can be defined as:
\begin{equation}
    \emph{X} = \textit{pairwise\_fusion}\left(\dot{\mathcal{G}},\dot{\emph{X}};\Theta_{p}\right)
    ~\label{eq:pw_fus}
\end{equation}
where $\Theta_{p}$ is the trainable parameters of any GNN model, such as GAT\cite{GAT} and PC-HGN\cite{PC-HGN}. The output $\emph{X} = \left[\emph{x}_{1},\emph{x}_{2},...,\emph{x}_{N}\right]\in\mathcal{R}^{d\times N}$ is the generated initial node embeddings that contain pairwise information. The effectiveness of pairwise fusion process will be further analyzed in the experiment section.
\subsection{Heterogeneous Hyperedge Generation}
The special structure of hyperedges enables hypergraphs to encode high-order data relations. By encircling a specific set of nodes with common attributes or close relationships within one hyperedge, we can effectively represent local group information among the nodes. Therefore, it is essential to discover the related nodes for different hyperedges. To facilitate hyperedge construction, we define two types of nodes: master node and slave node.  
A master node $v^{m}_{i}\in\mathcal{V}$ serves as an anchor when constructing a hyperedge, which dynamically encircles a set of slave nodes $\{v^{s}_i\}$ to jointly form the hyperedge. The collection of participant nodes is denoted as $S\left(v^{m}_{i}\right)$, which is united with master node $v_{i}^{m}$ to represent a complete data relation within a hyperedge. 
Previous works have adopted the $k$-NN method to select the \textit{k}-th nearest nodes to form a hyperedge~\cite{DHSL,huang2011unsupervised}. \revise{Alternative approaches solve an independent optimization problem to generate the hyperedges~\cite{wang2015visual,jin2019robust}, however, this solution fails to incorporate the heterogeneity of the hyperedges into the integrated optimization process.}

To address the above challenges, here we propose a dynamic learning process that generates heterogeneous hyperedges and adaptively reconstructs the hyperedge in different types. Specifically, for each master node $v^m_i\in\mathcal{V}$, our framework encloses the related slave nodes of various types to encode heterogeneous topological structures with implicit relationships, based on different attributes among the nodes. 
Suppose the master node $v^{m}_{i}$ has a set of $n_{i}$ candidate slave nodes $\Tilde{S}^{k}\left(v^{m}_{i}\right)$, where $k = \left[1,...,n_{i}\right]$ with different types. The hyperedge can be reconstructed based on the node types of the master node and candidate slave node sets.
The type of the candidate slave node set can be defined as $\mathcal{T}\left(\Tilde{S}^{k}\left(v^{m}_{i}\right)\right)$ and clearly is a subset of node type $\mathcal{T}_{v}$. Each candidate slave node set $\Tilde{S}^{k}\left(v^{m}_{i}\right)$ contains all nodes with one or several node types, which is denoted as $\Tilde{S}^{k}\left(v^{m}_{i}\right) = \left\{v|\phi\left(v\right)\in\mathcal{T}\left(\Tilde{S}^{k}\left(v^{m}_{i}\right)\right), v \neq v_{i}^{m}\right\}$. Since each hyperedge is generated from the corresponding candidate slave node set of a certain type, the type number of the hyperedge is equal to the type number of the candidate slave node set. \rev{An exemplary hyperedge construction is demonstrated in the middle part of \figurename~\ref{fig:overview}, wherein the master node encompasses $|\mathcal{T}_{v}|$ candidate slave node sets with distinct types, each of which is composed of all nodes with one single type. For each type, a hyperedge (represented as the coloured triangle) is dynamically constructed.}
Mathematically, this process of generating a hyperedge from a master node $v_{i}^{m}$ and one of its candidate slave node set $\Tilde{S}^{k}\left({v_i^{m}}\right)$ can be denoted as:
\begin{equation}
    \hat{\emph{X}}_{k}\left(v^{m}_{i}\right) = \emph{X}\left(\Tilde{S}^{k}\left(v^{m}_{i}\right)\right)\cdot\emph{p}_{v^{m}_{i}}^{k}\label{eq:recons}
\end{equation}
\vspace{-2mm}

It can be interpreted as the selection of slave nodes for this hypergraph through a linear combination of candidate slave node embedding $\emph{X}\left(\Tilde{S}^{k}\left(v^{m}_{i}\right)\right)\in\mathcal{R}^{d \times|\Tilde{S}^{k}\left(v^{m}_{i}\right)| }$ and trainable reconstruction coefficient vector $\emph{p}_{v^{m}_{i}}^{k}\in\mathcal{R}^{|\Tilde{S}^{k}\left(v^{m}_{i}\right)|}$. Each element $p_{v^{m}_{i}}^{k}\left(v\right)$ is the learned reconstruction coefficient associated with the node $v\in\Tilde{S}^{k}\left(v^{m}_{i}\right)$. Based on $\emph{p}_{v^{m}_{i}}^{k}$, the nodes in the candidate slave node set $\Tilde{S}^{k}\left(v^{m}_{i}\right)$ with reconstruction coefficient larger than zero are grouped to generate a hyperedge with the master node, which is denoted as $S^{k}\left(v^{m}_{i}\right)= \left\{v|v\in\Tilde{S}^{k}\left(v^{m}_{i}\right),\emph{p}_{v^{m}_{i}}^{k}\left(v\right)> 0\right\}$. The reconstruction error $c^{k}\left(v^{m}_{i}\right)$, which measures the difference between the master node and the reconstructed master node, is calculated as follows: 
\begin{equation}
    c^{k}\left(v^{m}_{i}\right) = \|\hat{\emph{X}}_{k}\left(v^{m}_{i}\right)-\theta_{k}\cdot\emph{X}\left(v^m_{i}\right)\|_{2}
    \label{eq:diff}
\end{equation}

$\theta_{k}\in\mathcal{R}^{d\times d}$ is a type-specific trainable projection matrix and $\|\cdot\|_{2}$ is the $l_{2}\text{-}norm$ regularization of a vector. Note that the purpose of hyperedge construction is to dynamically group the most relevant nodes, forming a hyperedge in which the nodes have strong dependencies. Practically, we add $l_{1}\text{-}norm$ regularization and $l_{2}\text{-}norm$ regularization of $\emph{p}$ simultaneously. The $l_{1}\text{-}norm$ regularization constrains the reconstruction coefficient to zero, which tends to group fewer slave nodes to reconstruct the master node. However, the $l_{1}\text{-}norm$ regularization is sensitive to noise and outliers, which leads to non-smooth results. For this reason, we consider the $l_{1}\text{-}norm$ regularization and $l_{2}\text{-}norm$ regularization simultaneously and use a weight hyperparameter $\gamma$ to trade off the effect of both. Overall, the loss of this process is defined as:
\begin{equation}
    \mathcal{L}_{recon} = \sum_{i = \left[1,...,N\right]}\sum_{k = \left[1,...,n_{i}\right]}\lambda c^{k}\left(v^{m}_{i}\right)+\|\emph{p}_{v^{m}_{i}}^{k}\|_{1}+\gamma\|\emph{p}_{v^{m}_{i}}^{k}\|_{2}
    ~\label{eq:lambda}
\end{equation}
where $\|\cdot\|_{1}$ denotes the $l_{1}\text{-}norm$ regularization of a vector, $\lambda$ is the weight hyperparameter of the reconstruction error, and $\gamma$ is the norm hyperparameter to trade off the $l_{1}\text{-}norm$ regularization and  $l_{2}\text{-}norm$ regularization of the reconstruction coefficient vector $\emph{p}_{v^{m}_{i}}^{k}$.

From the $n_{i}$ candidate slave node sets of the master node $v_{i}^{m}$, we thus can generate the $n_{i}$ hyperedges of different types. Iteratively, each node $v_{j}$ could act as the master node to generate several hyperedges of different types to lead to the final hypergraph. The corresponding incidence matrix $\emph{H}$ of the hypergraph is denoted as: 
\begin{equation}
\begin{aligned}
\emph{H}\left(v_{j},e^k\left(v_{i}^{m}\right)\right)&=
&\begin{cases}
1& \text{$v_{j} = v_{i}^{m}$}\\
\emph{p}_{v^{m}_{i}}^{k}\left(v_{j}\right)& \text{$v_{j}\in e^{k}\left(v^{m}_{i}\right),v_{j}\neq v_{i}^{m}$}\\
0& \text{otherwise}
\end{cases}
\end{aligned}
\label{eq:h}
\end{equation}

\vspace{-2mm}
\begin{figure*}[tb!]
    \centering
    \includegraphics[width=0.8\linewidth]{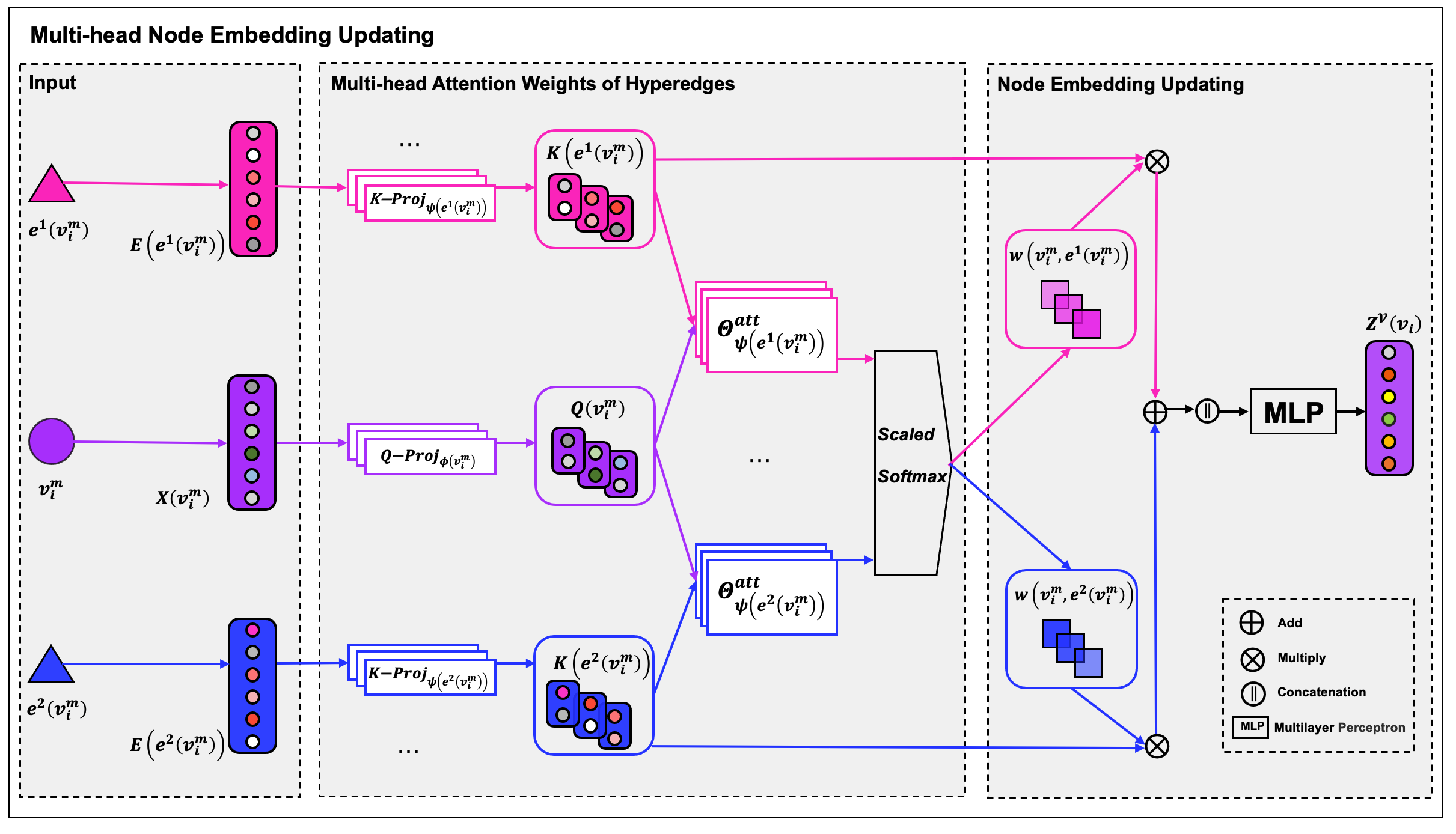}
    \vspace{-2mm}
  \caption{Type-specific multi-head attention on node embedding update.}
  \label{fig:4} 
  \vspace{-3mm}
\end{figure*} 

\subsection{Dynamic Hypergraph Learning}
As illustrated in Fig~\ref{fig:overview}, dynamic hypergraph learning consists of two key components: hyperedge embedding updating and multi-head attention node embedding updating. This section elaborates on these components and loss function.
\subsubsection{Hyperedge Embedding Updating}
We aggregate node embeddings to the hyperedge containing these nodes. The hyperedges embedding is denoted as $\emph{E}\in\mathcal{R}^{d\times M}$. The embedding of each hyperedge $e^{k}\left(v^m_{i}\right)$ is obtained as follows:
\begin{equation}
    \emph{E}\left(e^{k}\left(v^m_{i}\right)\right) = \frac{\emph{X}\cdot\emph{H}\left(e^{k}\left(v^m_{i}\right)\right)}{\delta\left(e^{k}\left(v^m_{i}\right)\right)}
    \label{eq:hyperedge}
\end{equation}
$\emph{H}\left(e^{k}\left(v^m_{i}\right)\right)\in\mathcal{R}^{N\times 1}$ is a column of incidence matrix $\emph{H}$ (See Eq.~\ref{eq:h}), where each element is a coefficient that represents the importance of the node to the hyperedge $e^{k}\left(v^m_{i}\right)$. The degree of edge $\delta\left(e^{k}\left(v^m_{i}\right)\right)$ is the normalization factor in the process of hyperedge embedding updating.

\subsubsection{Multi-head Attention Node Embedding Updating}
Previous works calculate the similarities among hyperedges and then use them as weighting coefficients when updating respective node embeddings~\cite{zhu2016heterogeneous,jin2019robust}. However, the heterogeneity in the hypergraph is omitted, especially with multiple types of hyperedge and nodes. To cope with that, we design a heterogeneous multi-head attention mechanism that can train the importance of heterogeneous hyperedges with respect to nodes as shown in \figurename~\ref{fig:4}. 

We use all hyperedges associated with master node $v^{m}_i$ to update the node embedding. The hyperedges associated with node $v^{m}_i$ is represented as $\mathcal{E}\left(v^m_i\right) = \left\{e^{1}\left(v^{m}_{i}\right),e^{2}\left(v^{m}_{i}\right),...,e^{M_{i}}\left(v^{m}_{i}\right)\right\}$.



\figurename~\ref{fig:4} illustrates the calculation process of a master node $v^{m}_i$ and two related hyperedges $e^{1}\left(v^{m}_{i}\right)$ and  $e^{2}\left(v^{m}_{i}\right)$: we first calculate the multi-head attention between $v_{i}^{m}$ and $e^{1}\left(v^{m}_{i}\right)$, and then use the normalized attention as the weight of $e^{1}\left(v^{m}_{i}\right)$. The calculation involves Eq.\ref{att} to Eq.\ref{att_2}.
\begin{equation}
    \emph{Q}^{h}\left(v^{m}_{i}\right) = Q\text{-}Proj^{h}_{\phi\left(v^m_i\right)}\cdot\emph{X}\left(v^m_{i}\right) \label{att}
\end{equation}
\rev{where the dimension of $\emph{X}\left(v^m_{i}\right)$ is $\mathcal{R}^{d\times 1}$, and $Q\text{-}Proj^{h}_{\phi\left(v^m_i\right)}\in\mathcal{R}^{\frac{d}{H}\times d}$ represents the projection matrix on the $h$-th attention head, $h\in\left[1,H\right]$. $H$ is the number of attention heads.} The master node is projected to generate the $h$-th query vector $\emph{Q}^{h}\left(v^{m}_{i}\right)$ by the projection matrix $Q\text{-}Proj^{h}_{\phi\left(v^m_i\right)}$. Note the projection matrices are distinguished on hyperedge types. Similarly, we also project the hyperedge $e^{1}\left(v^{m}_{i}\right)$ through $K\text{-}Proj^{h}_{\psi\left(e^{1}\left(v^{m}_{i}\right)\right)}$ into  $ \emph{K}^{h}\left(e^{1}\left(v^{m}_i\right)\right)$ as:

\begin{equation}
    \emph{K}^{h}\left(e^{1}\left(v^{m}_i\right)\right) = K\text{-}Proj^{h}_{\psi\left(e^{1}\left(v^{m}_{i}\right)\right)}\cdot\emph{E}\left(e^{1}\left(v^{m}_{i}\right)\right)
\end{equation}
where $\emph{E}\left(e^{1}\left(v^{m}_{i}\right)\right)\in\mathcal{R}^{d\times 1}$ is the embedding of hyperedge $e^{1}\left(v^{m}_{i}\right)$. The dimension of
\rev{$K\text{-}Proj^{h}_{\psi\left(e^{1}\left(v^{m}_{i}\right)\right)}$ is $\mathcal{R}^{\frac{d}{H}\times d}$} while the dimension of the output \rev{$\emph{K}^{h}\left(e^{1}\left(v^{m}_i\right)\right)$ is $\mathcal{R}^{\frac{d}{H}}$}.

\begin{equation}
    \begin{split}
    \label{att_1}
    att^{h}(v^{m}_{i},e^{1}(v^{m}_{i})) = \big(\emph{Q}^{h}&(v^{m}_{i})\cdot\Theta^{att}_{\psi\left(e^{1}\left(v^{m}_{i}\right)\right)}\\
    &\cdot\emph{K}^{h}\left(e^{1}\left(v^{m}_{i}\right)\right)\big)\cdot\frac{\mu_{\psi(e^{1}(v^{m}_{i}))}}{\sqrt{d}}
    \end{split}
\end{equation}
Eq. \ref{att_1} derives the attention value of the $h$-th head between $v^{m}_{i}$ and $e^{1}\left(v^{m}_{i}\right)$. The type-specific learnable matrix \rev{$\Theta^{att}_{\psi\left(e^{1}\left(v^{m}_{i}\right)\right)}\in\mathcal{R}^{\frac{d}{H}\times\frac{d}{H}}$} for edge type $\psi\left(e^{1}\left(v^{m}_{i}\right)\right)$ represents the learnable semantic information for each edge type. $\mu$ is a scaling factor for different hyperedge types. Moreover, since the magnitude of $\emph{K}$ and $\emph{Q}$ can increase the attention value significantly and eventually lead to the gradient explosion problem, we divide the obtained value by $\sqrt{d}$ to stabilize the training process.

Following that, the calculated attention value $att^{h}\left(v^{m}_{i},e^{1}\left(v^{m}_{i}\right)\right)$ is able to quantify the importance of $e^{1}\left(v^{m}_{i}\right)$ to $v^{m}_{i}$ on the $h$-th adaptively. The weight of hyperedge $e^{1}\left(v^{m}_{i}\right)$ is obtained as follows:
\begin{equation}
    \Tilde{\emph{w}}\left(v^{m}_{i},e^{1}\left(v^{m}_{i}\right)\right) = \mathop{\concat}\limits_{h\in\left[1,H\right]}att^{h}\left(v^{m}_{i},e^{1}\left(v^{m}_{i}\right)\right)
\end{equation}
where $\concat$ represents concatenation. The dimension of weight vector $\Tilde{\emph{w}}\left(v^{m}_{i},e^{1}\left(v^{m}_{i}\right)\right)$ is \rev{$\mathcal{R}^{H}$}, stacked by attention value on each head. To yield the final weight vector for each hyperedge, we gather the weight vectors from all related hyperedges of $\mathcal{E}\left(v^{m}_{i}\right)$ and conduct $Softmax$ as:

\begin{equation}
    \emph{w}\left(v^{m}_{i},e\right) = \mathop{Softmax}\limits _{\forall e\in\mathcal{E}\left(v^{m}_{i}\right)}\left(\Tilde{\emph{w}}\left(v^{m}_{i},e\right)\right)\label{att_2}
\end{equation}
where \rev{$\emph{w}\left(v^{m}_{i},e\right) = \left[\emph{w}\left(v^{m}_{i},e^{1}\left(v^{m}_{i}\right)\right),...,\emph{w}\left(v^{m}_{i},e^{M_{i}}\left(v^{m}_{i}\right)\right)\right]\in\mathcal{R}^{H\times M_{i}}$}, with entries are normalized among all hyperedges per head (illustrated in the right part of Fig.~\ref{fig:4}). The attentive aggregation of head $h$ for node $v^m_{i}$ is formulated as: 
\begin{equation}
    \emph{z}_{i}^{h} = \sum_{k\in\left[1,M_{i}\right]}\left(w^{h}\left(v^{m}_{i},e^{k}\left(v^m_i\right)\right)\times\emph{K}^{h}\left(e^{k}\left(v^m_{i}\right)\right)\right)
    \label{eq:node}
\end{equation}

$w^{h}\left(v^{m}_{i},e^{k}\left(v^m_i\right)\right)$ is the scalar element of $h$-th head in the weight vector $\emph{w}\left(v^{m}_{i},e^{k}\left(v^{m}_{i}\right)\right)$ of hyeredge $e^{k}\left(v^{m}_{i}\right)$, \rev{and the dimension of $\emph{z}_{i}^{h}$ is $\mathcal{R}^{\frac{d}{H}}$. After concatenating all $H$ heads,} we obtain the updated node embedding of $v^m_{i}$ through a shallow multi-layer perceptron (MLP), which is denoted as:
\begin{equation}
    \emph{Z}^{\mathcal{V}}\left(v^m_{i}\right) = \textit{MLP}\left(\mathop{\concat}\limits_{h\in\left[1,H\right]}\emph{z}^{h}_{i}\right)
    \label{eq:final_node}
\end{equation} 
\subsection{Loss Function}
We realize \pname in an end-to-end fashion, including the initialization of node embedding, the dynamic heterogeneous hyperedges generation and heterogeneous hyperedges learning. The learnt node representations can then be used in downstream tasks such as node classification and link prediction. The integrated training process is conducted in a supervised manner. We propose a united loss function, which combines the loss of reconstruction in the dynamic heterogeneous hyperedges generation and the loss of supervised downstream task. We take node classification as an example of a downstream task. In this case, the label of the node set $\mathcal{V}$ is $\emph{Y}^{\mathcal{V}}$. We use an MLP to map node representations to label space, and the loss in node classification can be calculated as: 
\begin{equation}
    \mathcal{L}_{label} = \textit{CE}\left(\textit{MLP}\left(\emph{Z}^{\mathcal{V}}\right),\emph{Y}^{\mathcal{V}}\right)
    \label{eq:label_loss}
\end{equation}
where $\textit{CE}$ is the cross-entropy function and used to measure the differences between estimated labels and the true labels of nodes. Finally, the united loss is defined as:

\begin{equation}
    \mathcal{L}_{u} = \left(1-\alpha\right)\times \mathcal{L}_{label}+\alpha \mathcal{L}_{recon}
    \label{eq:uloss}
\end{equation}
where $\mathcal{L}_{recon}$ is defined in Eq. \ref{eq:lambda}. The process of reconstructing the master node using slave nodes can uncover implicit relationships between the master node and others. However, as pointed out above, this process is sensitive to noisy nodes. Over-reliance on
reconstruction loss compromises the performance of our model. To enhance the generalization ability, we need to use a weight hyperparameter $\alpha$, to balance the effects of loss of supervised downstream task and reconstruction loss. 
The full process of \pname is described in Algorithm~\ref{lfh}.

\begin{algorithm}
\caption{Pseudocode of \pname training process \label{lfh}}
\begin{algorithmic}[1]
\REQUIRE The heterogeneous pairwise graph $\dot{\mathcal{G}}$; the raw node feature $\dot{\emph{X}}$; the set of node $\mathcal{V}$; the set of node type $\mathcal{T}_{v}$; \rev{the number of attention heads $H$;} the weight hyperparameter $\alpha$
\ENSURE Embedding of all nodes $\emph{Z}^{\mathcal{V}}$
\STATE $\emph{X}$ $\leftarrow$ Generate initial node embeddings by pairwise fusion (Eq.\ref{eq:pw_fus})
\FOR{each master node $v^{m}_{i} \in\mathcal{V}$}
\FOR{each node type $k \in\mathcal{T}_{v}$}
\STATE $\emph{X}(v^{m}_{i})$ $\leftarrow$ The embedding of the master node $v^{m}_{i}$
\STATE $\hat{\emph{X}_{k}}(v^{m}_{i})$ $\leftarrow$ Reconstruct $\emph{X}(v^{m}_{i})$ by the slave node set of the node type $k$ (Eq.\ref{eq:recons})
\STATE Compute the reconstruction error between $\emph{X}(v^{m}_i)$ and $\hat{\emph{X}_{k}}(v^{m}_{i})$ through Eq.\ref{eq:diff}
\ENDFOR
\ENDFOR
\STATE $\mathcal{L}_{recons}$ $\leftarrow$ Calculate the loss of reconstruction by Eq.\ref{eq:lambda}
\STATE $\emph{H}$ $\leftarrow$ Generate the incidence matrix of hypergraph by Eq.\ref{eq:h}
\FOR{each master node $v^{m}_{i} \in\mathcal{V}$}
\STATE $\mathcal{E}(v^{m}_{i})$ $\leftarrow$ Hyperedges embedding associated with $v^{m}_{i}$ updated by Eq.\ref{eq:hyperedge}
\ENDFOR
\FOR{each master node $v^{m}_{i}\in\mathcal{V}$}
\FOR{each $e(v^{m}_{i})\in\mathcal{E}(v^{m}_{i})$}
\FOR{each $h = 1,...,H$}
\STATE $w^{h}(v^{m}_{i},e(v^{m}_{i}))$ $\leftarrow$ Compute the normalized multi-head attention as the weight of the $h$-th head between $v^{m}_{i}$ and $e(v^{m}_{i})$ by Eq.\ref{att} to Eq.\ref{att_2}
\ENDFOR
\ENDFOR
\ENDFOR
\STATE $\emph{Z}^{\mathcal{V}}$ $\leftarrow$ Update node embedding from Eq.\ref{eq:node} to Eq.\ref{eq:final_node}
\STATE $\mathcal{L}_{label}$ $\leftarrow$ Compute the loss of supervised downstream task by Eq.\ref{eq:label_loss}.
\STATE Compute the united loss $\mathcal{L}_{u} = (1-\alpha)\times \mathcal{L}_{label}+\alpha\times\mathcal{L}_{recons}$ (Eq.\ref{eq:uloss}), and update all trainable weights.
\end{algorithmic}
\end{algorithm}

\vspace{-2mm}
\subsection{Model Analysis and Discussion}
We analyze the computational cost and summarize the advantages of \pname as below:
\subsubsection{Computational Cost}
In the process of heterogeneous hypergraph generation, the computational cost mainly lies in the reconstruction of the master node from its candidate slave node set (Eq.~\ref{eq:recons}). As it is essential to generate hyperedges of different types for each node, and the computational cost of heterogeneous hypergraph generation is $\emph{O}\left(d|\mathcal{T}_{v}|N\right)$, where $N$ is the number of nodes, $d$ is the dimensionality of node embedding, and $|\mathcal{T}_{v}|$ is the number of node types. Additionally, for the process of dynamic hypergraph learning, we use the incidence matrix $\emph{H}$ and nodes embedding $\emph{X}$ to update hyperedges embedding, with the computational cost of $\emph{O}\left(dMN\right)$. Note the number of hyperedges $M$ is proportional to the product of the $N$ and the number of edge types $|\mathcal{T}_{e}|$, thus the computational cost of hyperedges embedding updating is $\emph{O}\left(d|\mathcal{T}_{e}|N^2\right)$. Moreover, the computational cost of multi-head attention node embedding updating is $\emph{O}\left(d^{2}|\mathcal{T}_{e}|N\right)$, where $\mathcal{T}_{e}$ is the number of hyperedge types. Hence, the total computational cost of our proposal is $\emph{O}\left(d|\mathcal{T}_{e}|N^2+\left(d|\mathcal{T}_{v}|+d^2|\mathcal{T}_{e}|\right)N\right)$. The scale of $d$, $\mathcal{T}_{e}$ and $\mathcal{T}_{v}$ are usually far less than $N$, This computational cost is on par with many existing hypergraph-based models, such as \cite{wang2015visual} and \cite{yu2012adaptive}. 

\subsubsection{Discussion}
 Compared to pairwise graph learning models such as GAT \cite{GAT} and PC-HGN \cite{PC-HGN},  \pname is capable of modeling implicit high-order data relations. Moreover, \pname has two advantages. First, the process of hyperedges generation is conducted with hypergraph learning within a united training process while taking into account the heterogeneity of the graph. This enables an adaptive hypergraph modeling than HGNN \cite{FengAAAI19}, whose hyperedges are generated by clustering method \textit{k}-NN. Other works that use static clustering methods for hyperedge generation adopt different strategies of \textit{k}-means or a combination of \textit{k}-NN and \textit{k}-means to generate different hyperedges \cite{DHSL,6200340}. The main limitation of this line of work is the static clustering methods are sensitive to noise and outliers. Another issue is the hyperparameters of clustering method may affect the hyperedge generation, and adaptively calibrating the hyperparamters of clustering method during training is non-trivial. \pname, on the other hand, can incorporate hyperedges information of different types in different heads, using a heterogeneous multi-head attention mechanism. The mechanism dynamically quantifies the weights of hyperedges based on the correlation between the node representation and the related hyperedges representation, making the weights more descriptive of the importance of the related hyperedges to the node. 

\section{Experiment} ~\label{sec:experiment}

In this section, We conduct our experiments on three different graph datasets, including DBLP, IMDB, and ACM. We aim to answer the following research questions:
\begin{itemize}
    
    \item \textbf{RQ1:} How does our proposed \pname perform compared with the three kinds of baselines (homogenous graph learning, heterogeneous graph learning, and hypergraph learning models) for node classification task (\textbf{RQ1.1}) and link prediction task (\textbf{RQ1.2}).
    \item \textbf{RQ2:} What is the impact of hyperparameters $\lambda$ and $\gamma$ in hyperedge construction on the \pname performance?
    \item \textbf{RQ3:} How much do the node embedding size and \rev{the number of multi-head $H$} affect the final results?
    \item \textbf{RQ4:} How do different components, such as pairwise fusion, dynamic hypergraph construction and multi-head attention mechanism, affect the performance of \pname.
    \item \textbf{RQ5:} What is the impact of hyperparameter $\alpha$ on the unified loss?
\end{itemize}
\vspace{-6mm}

\subsection{Experimental Setup}\label{subsec:setting}

\begin{table}[]
\centering
 \caption{Statistics of the datasets.}
 \vspace{-2mm}
 \label{tab:datasets}
\scalebox{1}{
\begin{tabular}{l|lll}
\hline
Dataset          & ACM  & DBLP  & IMDB  \\ \hline
\# Nodes         & 8994 & 18405 & 12772 \\
\# Node Types    & 3    & 3     & 3      \\
\# Edges         & 12961     & 33973      & 18644      \\
\# Edge Types    & 4     & 4      & 4      \\
\# Node Features & 1902     & 334      & 1256      \\ \hline
\end{tabular}
}
\vspace{-2mm}
\end{table}
\vspace{-1mm}

\subsubsection{Dataset} The experiments are conducted on three datasets, which are popularly used in classification tasks. The details of these datasets are summarized in Table~\ref{tab:datasets}.
\begin{itemize}
    \item \textbf{DBLP}\footnote{https://s3.cn-north-1.amazonaws.com.cn/dgl-data/dataset/openhgnn/dblp4GTN.zip} is an  academic network from four research areas, including database, machine learning, data mining and information retrieval. It uses Paper (P), Author (A), and Conference (C) as different node types, while edges are presented as P-A, A-P, P-C and C-P with different edge types. Four research areas are used as labels for this dataset. The initial node features are calculated using bag-of-words. 
    \item \textbf{ACM}\footnote{ https://s3.cn-north-1.amazonaws.com.cn/dgl-data/dataset/openhgnn/acm4GTN.zip} shares a similar data characterises with DBLP. It contains Paper (P), Author (A) and Subject (S) as node types, along with four types of edges (P-A, A-P, P-S and S-P). The papers are labelled into three classes (Database, Wireless Communication, Data Mining). It also uses bag-of-words to construct the initial node features.
    \item \textbf{IMDB}\footnote{ https://s3.cn-north-1.amazonaws.com.cn/dgl-data/dataset/openhgnn/imdb4GTN.zip} contains \revises{Movie} (M), \revises{Actor} (A) and \revises{Director} (D). Each movie is labelled according to its genre (Action, Comedy, Drama). Node features are also initialized using bag-of-words.

\end{itemize}

\begin{table}[t!]
\caption{Shared parameter setup.}
\centering
{
\scalebox{1}{
\begin{tabular}{c||c}
\hline
Name & \ Setup\\
\hline
\hline
Optimizer & Adam~\cite{kingma2014adam}\\
Learning rate & 2e-3 \\
Dropout rate & 0.3 \\
Weight initializer & Xavier~\cite{glorot2010understanding}\\
Epoch & 100\\
MLP & 1 linear layer, 1 output layer \\
\hline
\end{tabular}}
}
\label{tab:hyperpara}
\vspace{-4mm}
\end{table}
\subsubsection{Baseline Models and Configurations} We compare with some state-of-the-art baselines, each of which reports promising results in the classification task. Although these baselines share the same goal of learning the representation of the graph, they were originally designed to fulfil graph learning on different graph data, including homogeneous pairwise graph learning, pairwise heterogeneous graph learning, and hypergraph learning.
\begin{itemize}
    \item \textbf{GCN}~\cite{GCN} is a graph convolutional network designed specifically for homogeneous pairwise graph learning. The depth of the layer in GCN is set to 2.
    \item \textbf{GAT}~\cite{GAT} is the first work that introduces the attention mechanism in homogeneous graph learning. It enables the weighted message aggregations from neighbour nodes. The number of attention heads is set to 3.
    \item \textbf{GraphSAGE}~\cite{hamilton2018inductive} designs a sampling approach when aggregating messages from neighbour nodes. It also supports different aggregation functions. The sample window of GraphSAGE is set to 10. 
    \item \textbf{GraphSAINT}~\cite{zeng2020graphsaint} splits nodes and edges from a bigger graph into a number of subgraphs, on which the GCN is applied for node representation learning. We adopt the node sampling strategy for GraphSAINT and use 8000 as the node budget and 25 as the number of subgraphs.
    \item \textbf{HAN}~\cite{WangWWW19} uses attention techniques on heterogeneous graph learning, in which the node embeddings are updated through manually designed meta-paths. The number of attention heads is set to 8.
    \item \textbf{HGT}~\cite{HuWWW20} designs type-specific attention layers, assigning different trainable parameters to each node and edge type. For the best performance of HGT, the number of attention heads is set to 8.
    \item \textbf{R-HGNN}~\cite{YuTKDE22} proposes a relation-aware model to learn the semantic representation of edges while discerning the node representations with respect to different relation types. It is designed to learn heterogeneous graph representations. The depth of the layer in R-HGNN is set to 2.  
    \item \textbf{PC-HGN}~\cite{PC-HGN} employs a sampling-based convolution. It designs an efficient cross-relation convolution that allows message aggregations of a node from different connected relation types simultaneously. We set the number of kernels to 64 and pooling size to 2 as the best performance reported.
    \item \textbf{HL}~\cite{HL} uses the hypergraphs to represent complex relationships among the objects, where where attribute values are regarded as hyperedges. The value of the regularization factor is set to 0.1.
    \item \textbf{HGNN}~\cite{FengAAAI19} performs convolution with a hypergraph Laplacian, which is further approximated by truncated Chebyshev polynomials to handle the data correlation during representation learning. The number of neighbors for generating hyperedges is set to 10.
    \item \textbf{HGNN+}~\cite{GaoTPAMI23} utilizes the multi-modal/multi-type correlation from each modality/type, and conducts hypergraph convolution in the spatial domain to learn a general data representation for various task. The value of $k$ for selecting $k$-hop neighbors to generate hyperedges is set to 1.
    \revise{\item \textbf{DHGNN}~\cite{Jiangijcai2019} utilizes the \textit{k}-NN method and \textit{k}-means clustering to dynamically construct the hypergraph, from which the hypergraph convolution is conducted for node representation learning. The number of clusters and nearest neighbors is set to 64.}
\end{itemize}

The shared parameter setup of the framework can be found in Table \ref{tab:hyperpara}. We randomly split all datasets into train/validate/test with the ratio of 0.2/0.1/0.7, respectively. It is worth mentioning that the hyperparameters of all baseline models are set to the optimal values reported in their original papers. All models are trained with a fixed 100 epochs, using an early stopping strategy when the performance on the validation set is not improved for 30 consecutive epochs. All trainable parameters of the neural network are initialized through Xavier \cite{glorot2010understanding} and optimized using Adam \cite{kingma2014adam} with the learning rate 2e-3. The dropout rate is set to be 0.3~\cite{srivastava2014dropout}. The optimal selection of hyperparameters of our model is further analysed in the following section.

\begin{table*}[t!]
\caption{Performance of different models on node classification task.}
\resizebox{\linewidth}{!}{
\begin{tabular}{cllllllllllllll}
\hline
\multicolumn{1}{l}{\textbf{Dataset}}                & \textbf{DSR} & \multicolumn{5}{c}{\textbf{Hypergraph-based}}                               & \multicolumn{4}{c}{\textbf{Heterogeneous-based}}                & \multicolumn{4}{c}{\textbf{Homogeous-based}}                           \\ \cmidrule(r){3-7} \cmidrule(r){8-11} 
                                            \cmidrule(r){12-15} &              &
\multicolumn{1}{l}{\textbf{LFH}} & \textbf{HGNN} & \textbf{HGNN+} & \textbf{HL} & \revise{\textbf{DHGNN}} & \textbf{PC-HCN} & \textbf{HAN} & \textbf{R-HGNN} & \textbf{HGT} & \textbf{GCN} & \textbf{GAT} & \textbf{GraphSAGE} & \textbf{GraphSAINT} \\ \hline
\multicolumn{1}{c|}{\multirow{4}{*}{\textbf{ACM}}}  & 10\%         &\multicolumn{1}{c}{\textbf{0.818}}              &\multicolumn{1}{c}{0.557}               &\multicolumn{1}{c}{0.655}                &\multicolumn{1}{c}{0.471}             &\multicolumn{1}{c}{\revise{0.486}}               &\multicolumn{1}{c}{0.768}                 &\multicolumn{1}{c}{0.551}              &\multicolumn{1}{c}{0.512}                 &\multicolumn{1}{c}{0.517}              &\multicolumn{1}{c}{0.505}              &\multicolumn{1}{c}{0.547}              &\multicolumn{1}{c}{0.548}                    &\multicolumn{1}{c}{0.512}                     \\
\multicolumn{1}{c|}{}                               & 30\%         &\multicolumn{1}{c}{\textbf{0.871}}              &\multicolumn{1}{c}{0.801}                &\multicolumn{1}{c}{0.851}                &\multicolumn{1}{c}{0.544}             &\multicolumn{1}{c}{\revise{0.513}}              &\multicolumn{1}{c}{0.845}                 &\multicolumn{1}{c}{0.842}              &\multicolumn{1}{c}{0.847}                 &\multicolumn{1}{c}{0.814}              &\multicolumn{1}{c}{0.762}              &\multicolumn{1}{c}{0.767}              &\multicolumn{1}{c}{0.763}                    &\multicolumn{1}{c}{0.771}                     \\
\multicolumn{1}{c|}{}                               & 50\%         &\multicolumn{1}{c}{\textbf{0.882}}              &\multicolumn{1}{c}{0.804}               &\multicolumn{1}{c}{0.852}                &\multicolumn{1}{c}{0.596}             &\multicolumn{1}{c}{\revise{0.781}}               &\multicolumn{1}{c}{0.849}                 &\multicolumn{1}{c}{0.842}              &\multicolumn{1}{c}{0.854}                 &\multicolumn{1}{c}{0.828}              &\multicolumn{1}{c}{0.846}              &\multicolumn{1}{c}{0.853}              &\multicolumn{1}{c}{0.851}                    &\multicolumn{1}{c}{0.839}                     \\
\multicolumn{1}{c|}{}                               & 70\%         &\multicolumn{1}{c}{\textbf{0.905}}              &\multicolumn{1}{c}{0.840}               &\multicolumn{1}{c}{0.882}                &\multicolumn{1}{c}{0.607}             &\multicolumn{1}{c}{\revise{0.881}}               &\multicolumn{1}{c}{0.872}                 &\multicolumn{1}{c}{0.866}              &\multicolumn{1}{c}{0.864}                 &\multicolumn{1}{c}{0.857}              &\multicolumn{1}{c}{0.853}              &\multicolumn{1}{c}{0.858}              &\multicolumn{1}{c}{0.858}                    &\multicolumn{1}{c}{0.856}                     \\ \hline
\multicolumn{1}{c|}{\multirow{4}{*}{\textbf{IMDB}}} & 10\%         &\multicolumn{1}{c}{\textbf{0.491}}              &\multicolumn{1}{c}{0.410}               &\multicolumn{1}{c}{0.432}                &\multicolumn{1}{c}{0.415}             &\multicolumn{1}{c}{\revise{0.417}}               &\multicolumn{1}{c}{0.407}                 &\multicolumn{1}{c}{0.372}              &\multicolumn{1}{c}{0.406}                 &\multicolumn{1}{c}{0.341}              &\multicolumn{1}{c}{0.322}              &\multicolumn{1}{c}{0.384}              &\multicolumn{1}{c}{0.351}                    &\multicolumn{1}{c}{0.361}                     \\
\multicolumn{1}{c|}{}                               & 30\%         &\multicolumn{1}{c}{\textbf{0.539}}              &\multicolumn{1}{c}{0.442}               &\multicolumn{1}{c}{0.463}                &\multicolumn{1}{c}{0.424}             &\multicolumn{1}{c}{\revise{0.440}}               &\multicolumn{1}{c}{0.459}                 &\multicolumn{1}{c}{0.398}              &\multicolumn{1}{c}{0.415}                 &\multicolumn{1}{c}{0.388}              &\multicolumn{1}{c}{0.384}              &\multicolumn{1}{c}{0.429}              &\multicolumn{1}{c}{0.388}                    &\multicolumn{1}{c}{0.409}                     \\
\multicolumn{1}{c|}{}                               & 50\%         &\multicolumn{1}{c}{\textbf{0.558}}              &\multicolumn{1}{c}{0.463}               &\multicolumn{1}{c}{0.446}                &\multicolumn{1}{c}{0.437}             &\multicolumn{1}{c}{\revise{0.471}}               &\multicolumn{1}{c}{0.461}                 &\multicolumn{1}{c}{0.427}              &\multicolumn{1}{c}{0.438}                 &\multicolumn{1}{c}{0.424}              &\multicolumn{1}{c}{0.438}              &\multicolumn{1}{c}{0.458}              &\multicolumn{1}{c}{0.427}                    &\multicolumn{1}{c}{0.422}                     \\
\multicolumn{1}{c|}{}                               & 70\%         &\multicolumn{1}{c}{\textbf{0.571}}              &\multicolumn{1}{c}{0.494}               &\multicolumn{1}{c}{0.492}                &\multicolumn{1}{c}{0.459}             &\multicolumn{1}{c}{\revise{0.528}}               &\multicolumn{1}{c}{0.488}                 &\multicolumn{1}{c}{0.464}              &\multicolumn{1}{c}{0.487}                 &\multicolumn{1}{c}{0.435}              &\multicolumn{1}{c}{0.445}              &\multicolumn{1}{c}{0.488}              &\multicolumn{1}{c}{0.454}                    &\multicolumn{1}{c}{0.460}                     \\ \hline
\multicolumn{1}{c|}{\multirow{4}{*}{\textbf{DBLP}}} & 10\%         &\multicolumn{1}{c}{\textbf{0.861}}              &\multicolumn{1}{c}{0.669}               &\multicolumn{1}{c}{0.819}                &\multicolumn{1}{c}{0.503}             &\multicolumn{1}{c}{\revise{0.481}}               &\multicolumn{1}{c}{0.818}                 &\multicolumn{1}{c}{0.812}              &\multicolumn{1}{c}{0.672}                 &\multicolumn{1}{c}{0.598}              &\multicolumn{1}{c}{0.535}              &\multicolumn{1}{c}{0.574}              &\multicolumn{1}{c}{0.543}                    &\multicolumn{1}{c}{0.597}                     \\
\multicolumn{1}{c|}{}                               & 30\%         &\multicolumn{1}{c}{\textbf{0.874}}              &\multicolumn{1}{c}{0.739}               &\multicolumn{1}{c}{0.842}                &\multicolumn{1}{c}{0.575}             &\multicolumn{1}{c}{\revise{0.512}}               &\multicolumn{1}{c}{0.838}                 &\multicolumn{1}{c}{0.837}              &\multicolumn{1}{c}{0.783}                 &\multicolumn{1}{c}{0.632}              &\multicolumn{1}{c}{0.699}              &\multicolumn{1}{c}{0.693}              &\multicolumn{1}{c}{0.696}                    &\multicolumn{1}{c}{0.676}                     \\
\multicolumn{1}{c|}{}                               & 50\%         &\multicolumn{1}{c}{\textbf{0.888}}              &\multicolumn{1}{c}{0.767}               &\multicolumn{1}{c}{0.866}                &\multicolumn{1}{c}{0.634}             &\multicolumn{1}{c}{\revise{0.751}}               &\multicolumn{1}{c}{0.861}                 &\multicolumn{1}{c}{0.855}              &\multicolumn{1}{c}{0.853}                 &\multicolumn{1}{c}{0.724}              &\multicolumn{1}{c}{0.755}              &\multicolumn{1}{c}{0.775}              &\multicolumn{1}{c}{0.771}                    &\multicolumn{1}{c}{0.776}                     \\
\multicolumn{1}{c|}{}                               & 70\%         &\multicolumn{1}{c}{\textbf{0.917}}              &\multicolumn{1}{c}{0.791}               &\multicolumn{1}{c}{0.892}                &\multicolumn{1}{c}{0.656}             &\multicolumn{1}{c}{\revise{0.784}}               &\multicolumn{1}{c}{0.884}                 &\multicolumn{1}{c}{0.878}              &\multicolumn{1}{c}{0.863}                 &\multicolumn{1}{c}{0.782}              &\multicolumn{1}{c}{0.772}              &\multicolumn{1}{c}{0.780}              &\multicolumn{1}{c}{0.778}                    &\multicolumn{1}{c}{0.778}                     \\ \hline
\end{tabular}
}

\label{tab:nc}
\end{table*}

\begin{figure}[t!]
\vspace{-3mm}
        \captionsetup{labelfont={color=black}}
        \centering
        \includegraphics[width=0.9\linewidth]{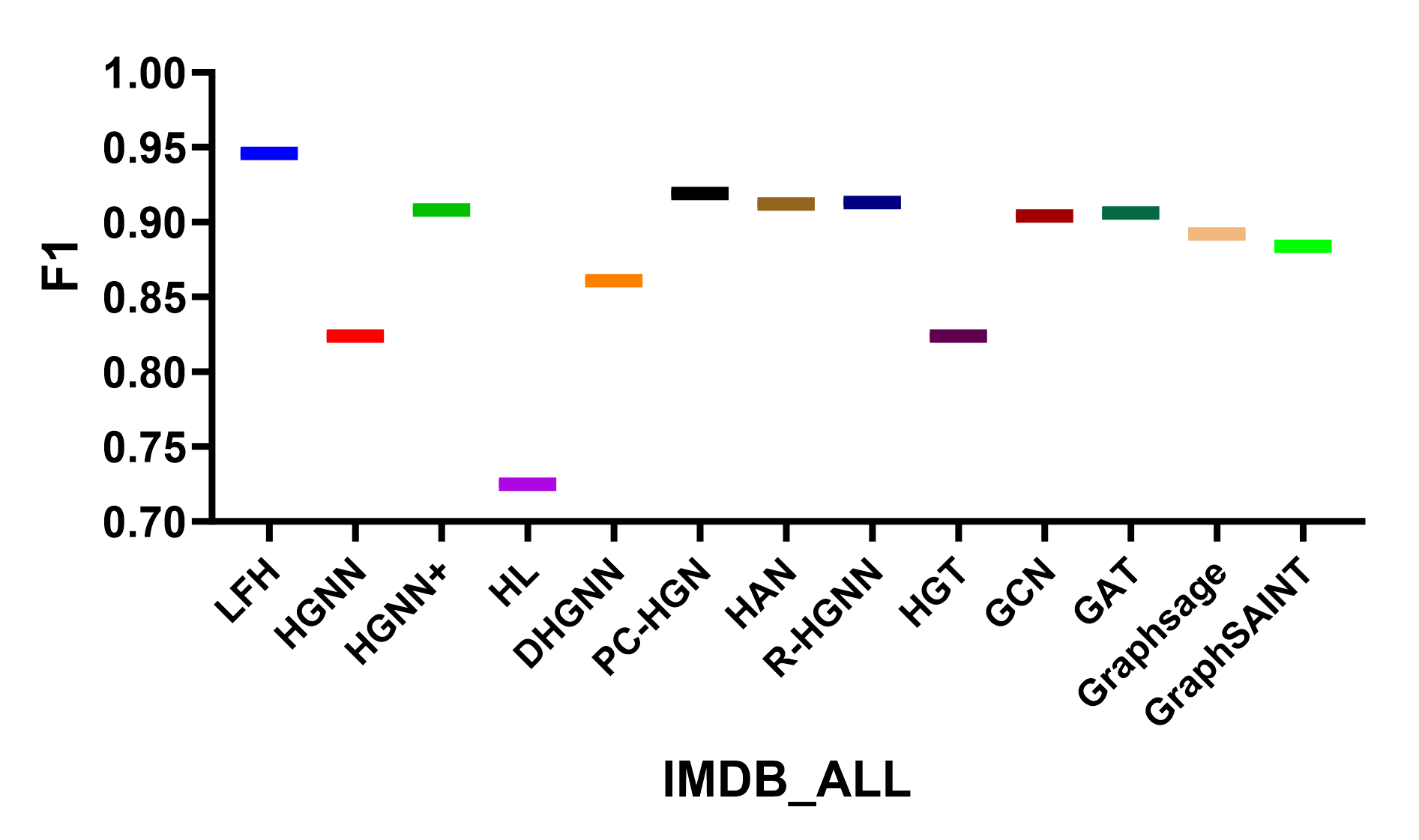}
        \vspace{-3mm}
		\caption{\rev{Comparison of F1 on IMDB\_ALL.}}
		\label{fig:scala}
		\vspace{-5mm}
\end{figure}

\vspace{-2mm}
\subsection{Performance Analysis (RQ1)}
We evaluate the performance of our proposed framework on node classification and link prediction tasks. Experiments for both tasks are
repeated three times, and the average metric is reported.
\subsubsection{Node Classification (RQ1.1)} \revise{In the node classification task, we initially perform a random sampling of nodes at varying percentages (10\%, 30\%, 50\%, 70\%) of the nodes from the three datasets (ACM, IMDB and DBLP). Then, any edges connected to nodes that are not sampled are eliminated. Subsequently, any sampled nodes that lack neighboring nodes are also removed. It is observed that a lower data sampling ratio correlates with an increased introduction of noise into the training dataset, due to the loss of original information.} 
The best F1 results among all datasets are demonstrated in Table~\ref{tab:nc}, in which the highest scores are marked in bold. We compared three different categories of graph learning models: homogeneous pairwise graph learning models, heterogeneous pairwise graph learning models, and hypergraph learning models. Our proposed \pname achieves performance gain over other baseline models in all datasets by 2\%-38\%. \revise{Our model outperforms all baseline models in all datasets by an average improvement of 12.9\%.} Compared to four homogeneous pairwise GNN baselines GCN, GAT, GraphSAGE and GraphSAINT, the average F1 score improvement of \pname for all datasets is 14.4\%. \revise{Regarding the cherry-picked heterogeneous pairwise graph learning models and hypergraph learning models, our model achieved an average improvement of 9.7\% and 14.6\%, respectively, compared to the best-recorded results in all datasets.} It is worth noting that HGNN+ \cite{GaoTPAMI23} is the only hypergraph learning baseline that utilizes pairwise information for hyperedge generation, resulting in better performance compared to other hypergraph learning models, except for our model. Furthermore, the training process of our model, which incorporates type-specific hyperedge generation and attentive embedding updates that exploits the heterogeneity of the graph, enables \pname to improve HGNN+ by an average of 5.7\% in all datasets. Additionally, compared to the latest reporting heterogeneous learning model PC-HGN \cite{PC-HGN}, the best results of our model gains an average increase by 5.2\%. \revise{In general, we observe that our model outperforms all baselines for different data sample ratios (DSR), which is shown in Table~\ref{tab:nc}. When the data sample ratios are set to be 70\%, 50\%, 30\% and 10\%, our model outperforms other models by 10.1\%, 10.2\%, 10.9\% and 16.7\% on average, respectively. It is worth noting that our model achieves higher average improvements when the data sampling ratio is set to 10\%, and then the margin of average improvements declines as the data sampling ratio increases.} 
\rev{To further verify the scalability of LFH, we conduct the experiment on the intact IMDB dataset, named IMDB\_ALL, which contains 18405 nodes and 33973 edges. The experimental results on the IMDB\_ALL demonstrate that LFH maintains superior performance over all other baselines, as shown in Fig.~\ref{fig:scala}, with an average performance gain of around 7.3\%.}

\begin{table}[t!]
 \caption{Performance of different models on link prediction.}
 \label{tab:pairwise}
 \centering
 \scalebox{1}{
 \begin{tabular}{c|ccc}
   \hline
   \bf{Model}  & \bf{ACM}  & \bf{DBLP}  & \bf{IMDB} \\
   \hline 
   \bf{Node2Vec \cite{node2vec}}&0.689&0.628&0.724\\
   \bf{GAT \cite{GAT}}&0.671&0.606&0.703\\
   \bf{PC-HGN \cite{PC-HGN}}&0.833&0.782&0.842\\
   \bf{HGNN+ \cite{GaoTPAMI23}}&0.632&0.689&0.784\\
   \bf{\revise{DHGNN \cite{Jiangijcai2019}}}&\revise{0.714}&\revise{0.735}&\revise{0.696} \\
   \bf{\pname}&\bf{0.861}&\bf{0.809}&\bf{0.874}\\
   \hline
  \end{tabular}
  }
\label{tab:lp}
\vspace{-4mm}
\end{table}
\subsubsection{Link Prediction (RQ1.2)} To investigate the effectiveness of our proposed framework, we also apply our model to the link prediction task. In this task, a graph with a certain faction of edges removed is given, and thus the objective is to predict these missing edges. Following the experiment setting adopted in \cite{node2vec}, 50\% percent of existing edges are randomly set hidden as positive samples while ensuring that the residual graph after the edge being removed remains connected. To generate negative examples, we randomly sample an equal number of node pairs from the graph, which have no edge connecting them. Similar to the related works~\cite{FanPAMI22,poincare}, we use a logistic regression classifier as a predictor for link prediction. To get the representation of each edge, we use Hadamard operator to compute the element-wise product for the linked node pairs. We compare the performance of \pname with five baselines. \revise{Node2vec \cite{node2vec} is a classic model for link prediction task. DHGNN is a hypergraph learning method that incorporates dynamic hypergraph construction. The other three baselines, GAT, PC-HGN, and HGNN+, are the top-performing baselines in homogenous pairwise graph learning models, heterogeneous pairwise graph learning models, and hypergraphs learning models in node classification tasks. Experimental results are summarized in Table~\ref{tab:lp}. \pname obtains performance gains of 16.8\%, 18.8\%, 2.9\%, 14.6\% and 13.7\%, compared with Node2vec, GAT, PC-HGN, HGNN+, DHGNN and outperforms five baseline models in three datasets by an average improvement of 12.8\%.} It shows the promising capability of our model to capture implicit data relations. It is worth noting that HGNN+ that performs well in node classification does not have comparable performance in link prediction. This is because the HGNN+ applied for link prediction task heavily relies on pairwise edges to generate hyperedges, and thus a sharp decrease in the number of existing edges leads to a deterioration in performance.

\begin{figure*}[t!]
    \vspace{-3mm}
    \centering
  \scalebox{0.85}{
  \subfloat[ACM\label{p1-acm}]{%
        \includegraphics[width=0.33\linewidth]{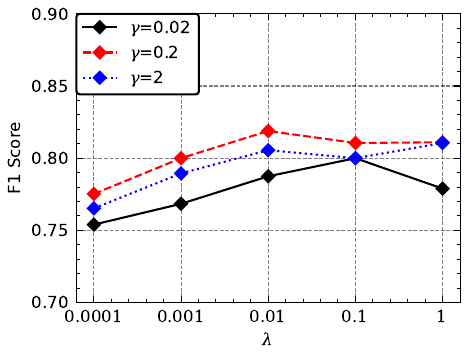}}
    \hfill
  \subfloat[DBLP\label{p1-dblp}]{%
        \includegraphics[width=0.33\linewidth]{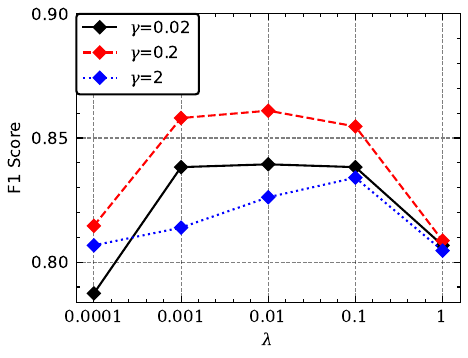}}
           \hfill
    \subfloat[IMDB\label{p1-imdb}]{%
       \includegraphics[width=0.33\linewidth]{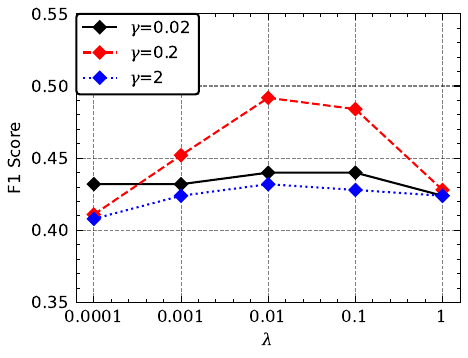}}
    }
  \caption{Performance of \pname as a function of $\lambda$ for several values of $\gamma$ for different datasets.}
  \vspace{-6mm}
  \label{fig:3g5l} 
\end{figure*}

\begin{figure*}[t!]
    \centering
    
  \scalebox{0.85}{
  \subfloat[ACM\label{p2-acm}]{%
        \includegraphics[width=0.33\linewidth]{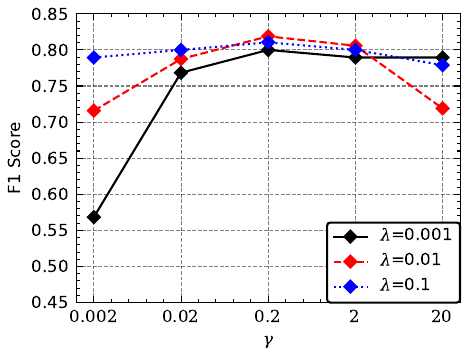}}
    \hfill
  \subfloat[DBLP\label{p2-dblp}]{%
        \includegraphics[width=0.33\linewidth]{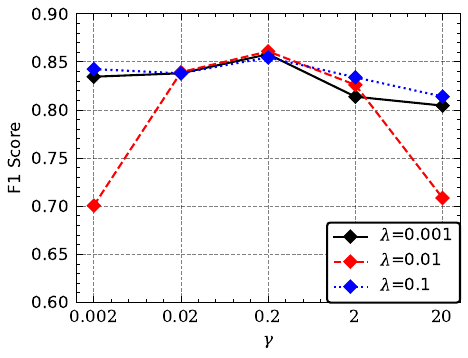}}
    \hfill
   \subfloat[IMDB\label{p2-imdb}]{%
       \includegraphics[width=0.33\linewidth]{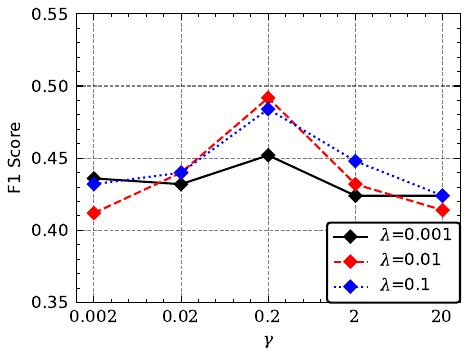}}
    }
  \caption{Performance of \pname as a function of $\gamma$ for several values of $\lambda$ for different datasets.}
  \label{fig:5g3l} 
  \vspace{-5mm}
\end{figure*} 

\vspace{-1mm}
\subsection{Impact of Hyperedge Construction (RQ2)}
In our proposed \pname, $\lambda$ and $\gamma$ are used as hyperparameters. Specifically, $\lambda$ is the weight hyperparameter of the reconstruction error and $\gamma$ is the norm hyperparameter to trade off $l_{1}\text{-}norm$ regularization norm and $l_{2}\text{-}norm$ regularization of the reconstruction coefficient vector (See Eq.~\ref{eq:lambda}). \figurename~\ref{fig:3g5l} and \ref{fig:5g3l} illustrate the overall performance of \pname influenced by $\lambda$ and $\gamma$. \figurename~\ref{fig:3g5l} illustrates the influence of $\lambda$ for different datasets on F1 score. In this figure, we fix $\gamma$ as 0.02, 0.2 and 0.2 while $\lambda$ ranges between 0.0001 and 1. When $\lambda$ is small, the reconstruction loss of $\lambda c^{k}\left(v^{m}_{i}\right)$ becomes trivial, and the reconstruction coefficient vector $\emph{p}$ will not experience large adjustments throughout the training phase, thus affecting the generation of hyperedges and deteriorating the performance of \pname. In contrast, when $\lambda$ increases to around 0.2, reconstruction loss of $\lambda c^{k}\left(v^{m}_{i}\right)$ accounts for the appropriate proportion of the total loss, thus achieving the best performance under this setup. When $\lambda$ continually increases to 20, F1 score drops rapidly, which clearly demonstrates that the dominance of reconstruction loss compromises the generalization ability of \pname.

On the other hand, \figurename~\ref{fig:5g3l} describes the influence of $\gamma$. In this figure, we fix $\lambda$ as 0.001, 0.01 and 0.1, while $\gamma$ ranges between 0.002 and 20. When $\gamma$ is very small, The decisive role of $l_{1}\text{-}norm$ regularization enables our model to encircle fewer slave nodes to generate hyperedges, which makes our model more sensitive to noise and outliers. When $\gamma$ is large, the dominance of $l_{2}\text{-}norm$ regularization leads to the over-smoothing problem. With $\gamma$ ranging in $\left[ 0.0001,1\right]$, the F1 score reaches the highest when $\gamma = 0.2$ for all fixed value of $\lambda$ in all datasets, and decreases afterwards. The result demonstrates that considering $l_{1}\text{-}norm$ regularization norm and $l_{2}\text{-}norm$ regularization simultaneously can improve the generalization ability of our model.

\vspace{-1mm}
\subsection{Sensitivity Analysis (RQ3)}
We study the impact of key parameters in \pname including the node embedding size and the number of attention head as illustrated in Table~\ref{tab:emb} and \ref{tab:head}.
\subsubsection{Impact of the node embedding size}
We test the impact of different node embedding sizes for all datasets with data split ratio of 10\%. As shown in Table~\ref{tab:emb}, the performance of the model increases at first and then
starts to drop, reporting the best result with the embedding
size 256. Intuitively, larger embedding sizes could lead to overfitting, 
thus causing unexpected performance drops.
\subsubsection{Impact of the number of muti-head}
We analyse the effect of \rev{the number of multi-head $H$.} As observed in Table~\ref{tab:head}, when the number of attention heads increases, there is typically an improvement in the performance of our model. However, it is observed that the performance of \pname improves only marginally afterwards while causing large computational costs. \rev{When $H = 4$}, it is an optimal point to trade off the performance and computational cost. 
\begin{table}[t!]
 \caption{Performance of \pname with different node embedding sizes.}
 \vspace{-3mm}
 \label{tab:emb}
 \centering
 \scalebox{1.0}{

 \begin{tabular}{c|ccc}
   \hline
   \bf{Initial Node Embedding Size}  & \bf{ACM}  & \bf{DBLP}  & \bf{IMDB} \\
   \hline 
   \bf{128}&0.800&0.818&0.456\\
   \bf{256}&0.818&0.861&0.491\\
   \bf{512}&0.789&0.766&0.424\\\hline
  \end{tabular}
  }
  \vspace{-1mm}
\end{table}

\begin{table}[t!]
 \caption{Performance of \pname with different number of attention heads.}
 \label{tab:head}
 \centering
 \scalebox{1.0}{

 \begin{tabular}{c|ccc}
   \hline
   \bf{Attention Head Number}  & \bf{ACM}  & \bf{DBLP}  & \bf{IMDB} \\
   \hline 
   \bf{2}&0.801&0.850&0.472\\
   \bf{4}&0.818&0.861&0.491\\
   \bf{8}&0.819&0.862&0.491\\\hline
  \end{tabular}
  }
\end{table}

\begin{figure}[t!]
        \centering
        \includegraphics[width=0.8\linewidth]{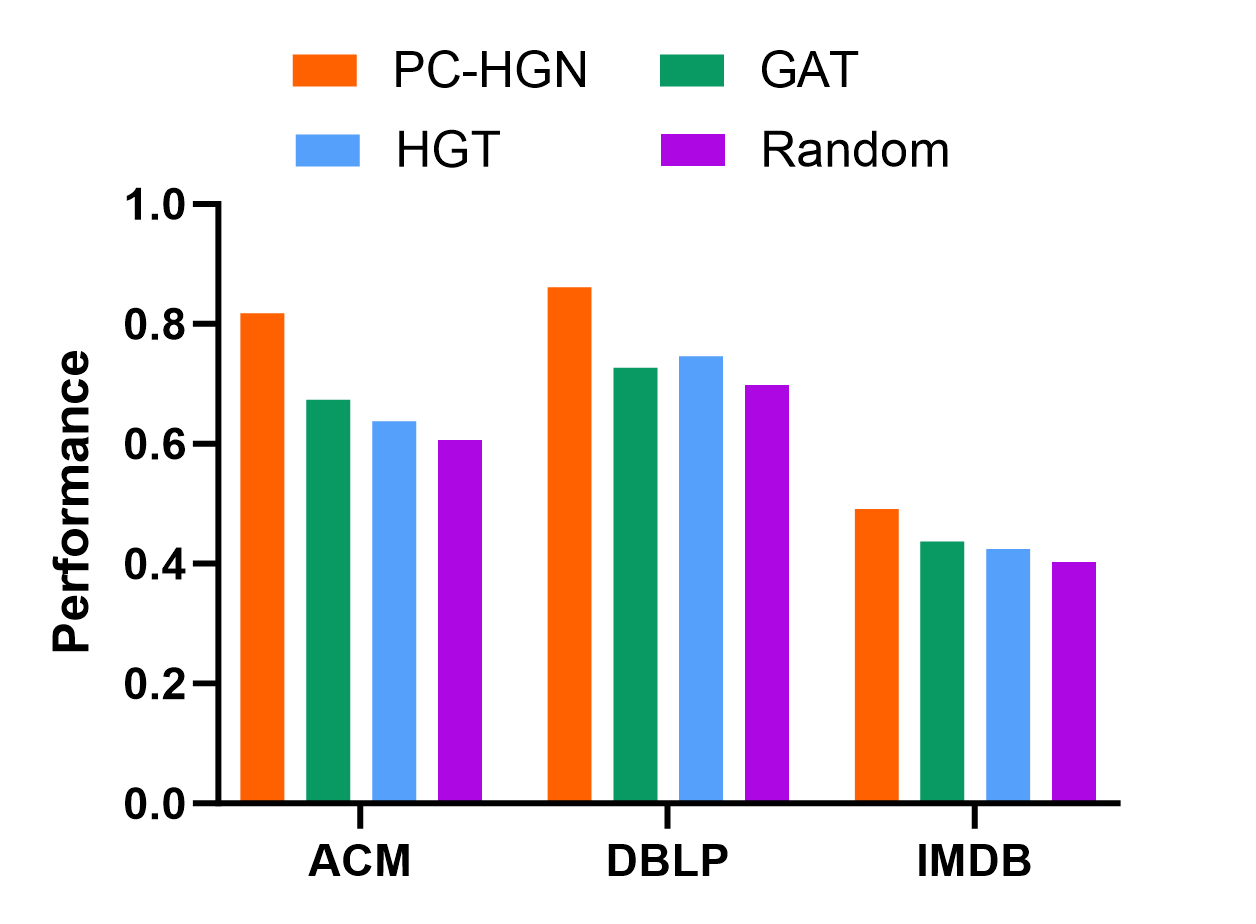}
          \vspace{-3mm}
		\caption{\rev{Ablation on using different GNN models and random initialization strategies for initial embedding generation.}}
		\label{fig:ablation}
		\vspace{-3mm}
\end{figure}

\begin{table}[]
\caption{Performance of LFH for ablation study.}
\centering
\begin{tabular}{l|lll}
\hline
\textbf{Model}                                & \textbf{ACM} & \textbf{DBLP} & \textbf{IMDB} \\ \hline
\textbf{LFH}                                  & 0.818        & 0.861         & 0.491         \\ \hline
\textbf{-w/o Dynamic Hypergraph Constrcution} & 0.774        & 0.673         & 0.405         \\ \hline
\textbf{-w/o Multi-head Attention}            & 0.653        & 0.582         & 0.344         \\ \hline
\end{tabular}
\vspace{-3mm}
\label{table:ablation}
\end{table}

\subsection{Ablation Study (RQ4)}
We study the impact of pairwise fusion in this part. We consider different candidates in the pairwise fusion process, aiming to find the best match that is capable of deriving high-quality features from the heterogeneous pairwise graph as the initial node embedding. We analyse the importance of pairwise fusion and implement different models introduced in Section~\ref{subsec:setting}. Note that we applied the same hyperparameters of all these models according to the best performance reported in the respective papers and source codes. \rev{As shown in \figurename~\ref{fig:ablation},  using PC-HGN in pairwise fusion to generate initial node embeddings significantly enhances performance across all datasets. It yields a performance gain of more than 5.4\% to 14.4\% compared to GAT, more than 6.6\% to 18.0\% compared to HGT, and an average performance gain of 10.5\% compared to the random initialization strategy. This result further verify the great importance to generate high-quality initial node embedding to better exploit the implicit data relations in the hypergraph. Additionally, it demonstrates the initial node embedding generated by PC-HGN helps our framework achieve the best results in all three datasets.}

In order to validate the effectiveness of different components in \pname, we design two variants of \pname, and compare the performance of these variants with \pname in the node classification task. The results of comparison are shown in Table~\ref{table:ablation}. Specifically, we investigate the contributions of dynamic hypergraph construction and multi-head attention to the performance of \pname. \pname w/o dynamic hypergraph construction replaces the learning process with unsupervised $k$-NN method for generating heterogeneous hyperedges, and the number of nearest nodes to form a hyperedge is set to 10~\cite{FengAAAI19}. The results show that \pname has a better performance over \pname w/o dynamic hypergraph construction by margins of 4.4\%-18.9\%, which indicates that the learning process for generating hypergraph enables \pname to encode topological structure more accurately. \pname w/o multi-head attention excludes the importance of heterogeneous hyperedges with respect to the master node, and assigns the weights to different heterogeneous hyperedges equally during hypergraph learning. The performance of \pname w/o multi-head attention drops significantly in all three datasets. The result demonstrates the significance of the dynamically learned weights of hyperedges and the effectiveness of the multi-head attention mechanism.

\begin{figure*}[t!]
  \vspace{-4mm}
    \centering
    \scalebox{0.8}{
  \subfloat[ACM\label{alpha-acm}]{%
        \includegraphics[width=0.33\linewidth]{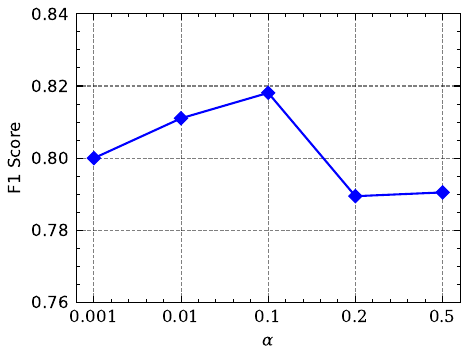}}
    \hfill
  \subfloat[DBLP\label{alpha-dblp}]{%
        \includegraphics[width=0.33\linewidth]{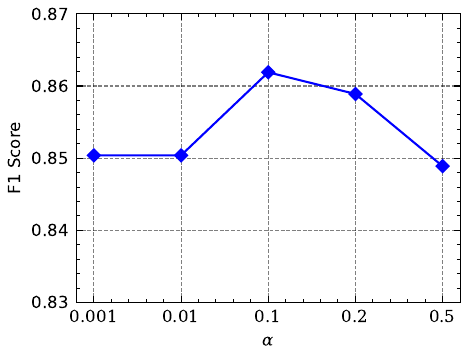}}
    \hfill
   \subfloat[IMDB\label{alpha-imdb}]{%
       \includegraphics[width=0.33\linewidth]{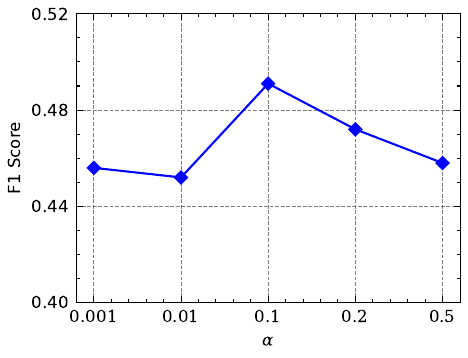}}
    }
  \caption{The performance change among different datasets with the
change of $\alpha$.}
  \label{fig:loss_alpha} 
  \vspace{-4mm}
\end{figure*} 

\subsection{Impact of Hyperparameter $\alpha$ (RQ5)}
As defined in Eq. \ref{eq:lambda}, the proposed united loss is a linear combination of the reconstruction loss in dynamic heterogeneous hyperedges generation and the supervised loss for the downstream task. We further study the impact of $\alpha$ on the performance of \pname. In the experiment, the sensitivity of $\alpha$ is explored to control the level of impact caused by the reconstruction of each master when training the model. When the value of $\alpha$ is close to 1, it means that the model is optimised towards the reconstruction correctness. Fig.~\ref{fig:loss_alpha} reveals the change of F1 score along with $\alpha$. With $\alpha$ ranging in $\left[0.001,0.5\right]$, the performance first reaches to the top and then drops significantly, indicating that over-reliance on reconstruction loss aggravates the generalization ability of our model. We observe that F1 score reaches the highest when $\alpha = 0.1$ in all three datasets. The result demonstrates that uniting reconstruction loss in the training process, to some extent, can help improve the model performance.

\begin{figure}[t!]
\vspace{-2mm}
        \captionsetup{labelfont={color=black}}
        \centering
        \includegraphics[width = 0.99\linewidth]{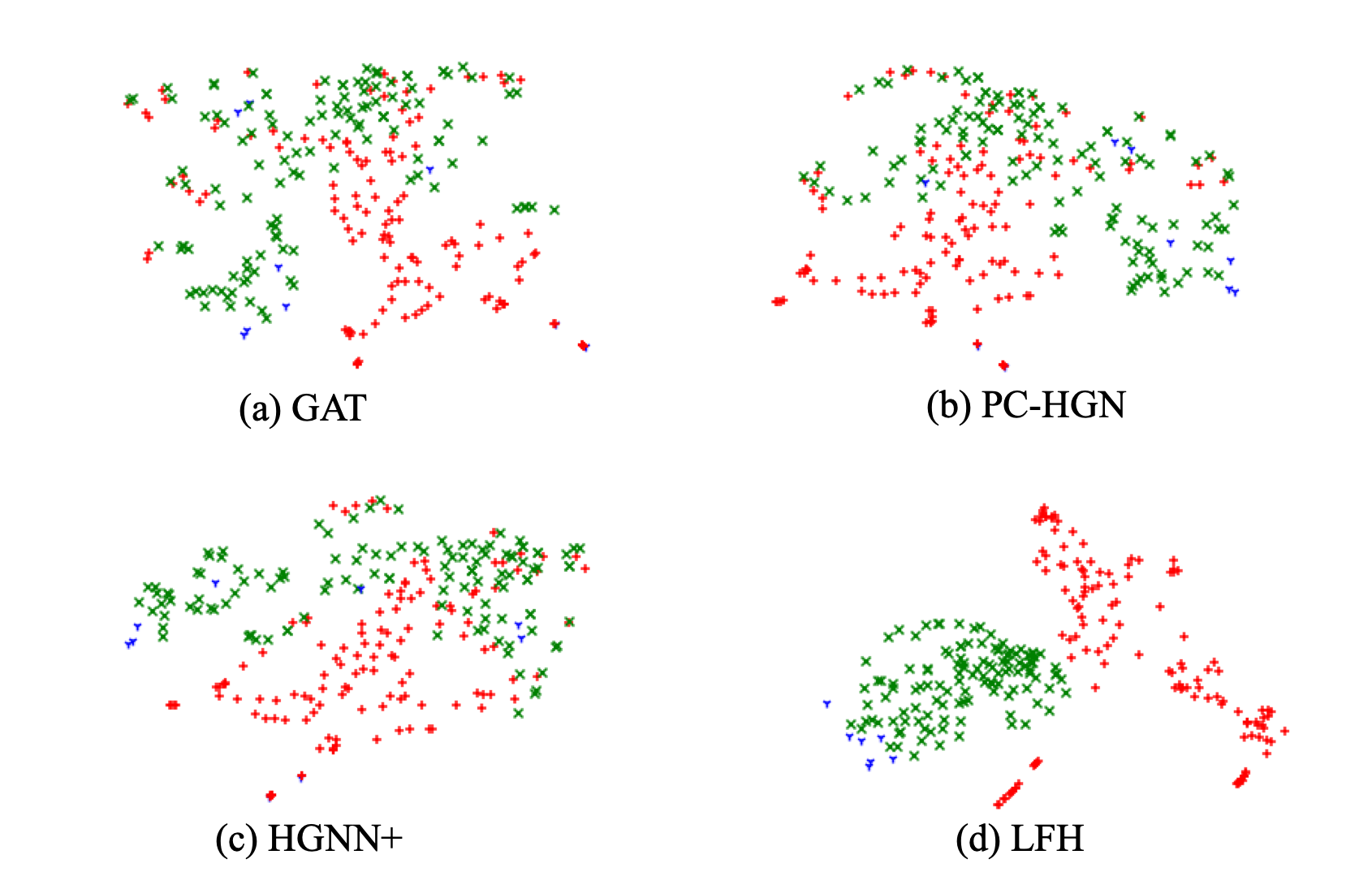}
        	\vspace{-4mm}
		\caption{\revises{A t-SNE visualization of our proposed model compared to baseline models on the ACM dataset (red indicates paper nodes, green denotes author nodes, and blue represents subject nodes).}}
		\label{fig:vis}
          \vspace{-4mm}

\end{figure}
\begin{figure}[t!]
        \captionsetup{labelfont={color=black}}
	\vspace{-2mm}
        \centering
        \includegraphics[width=0.99\linewidth]{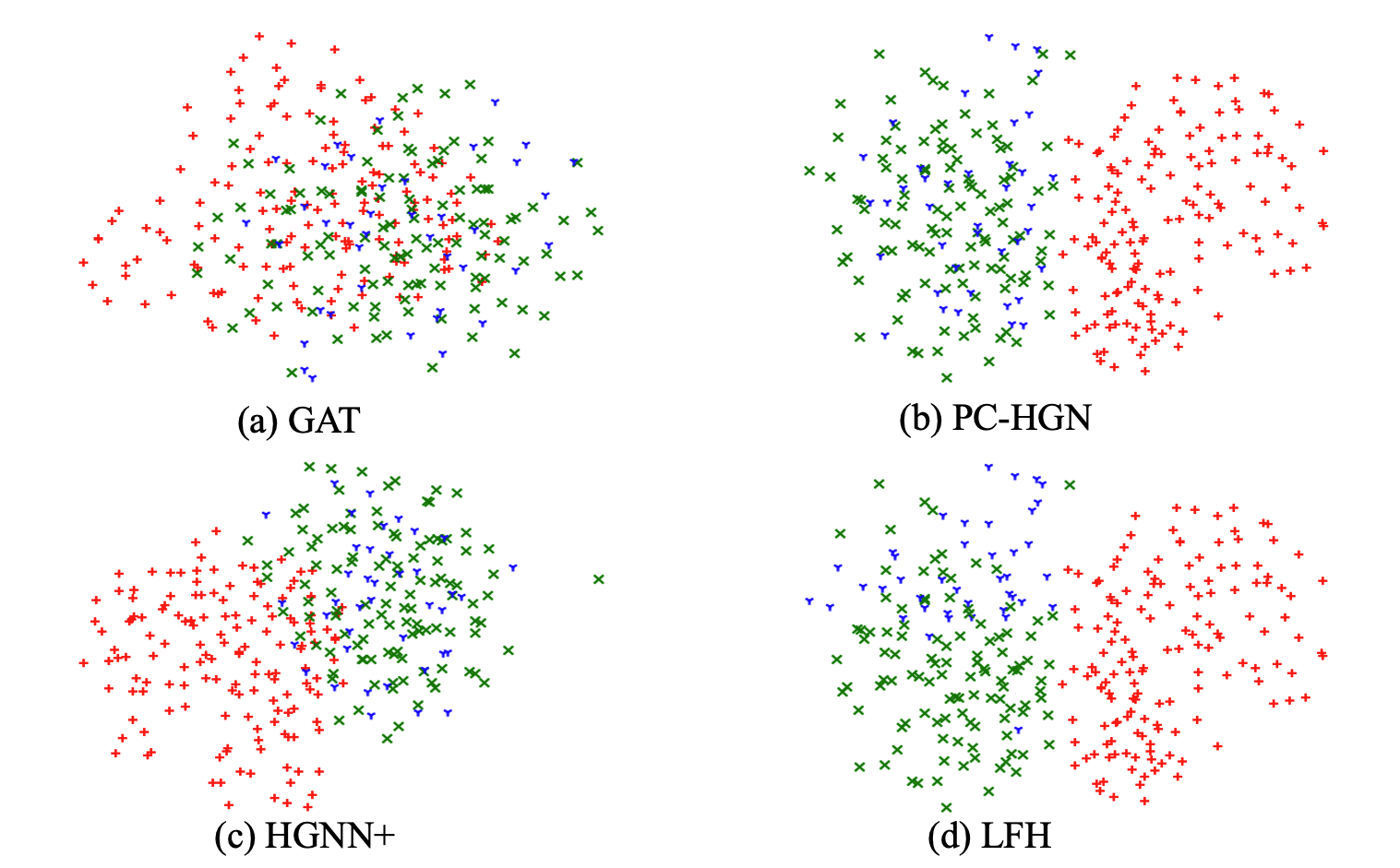}
        	\vspace{-5mm}
		\caption{\rev{A t-SNE visualization of our proposed model compared to baseline models on the IMDB dataset (red indicates actor nodes, green denotes movie nodes, and blue represents director nodes).}}
		\label{fig:vis_imdb}
		\vspace{-2mm}
\end{figure}
\begin{figure}[t!]
        \captionsetup{labelfont={color=black}}
	\vspace{-2mm}
        \centering
        \includegraphics[width=0.99\linewidth]{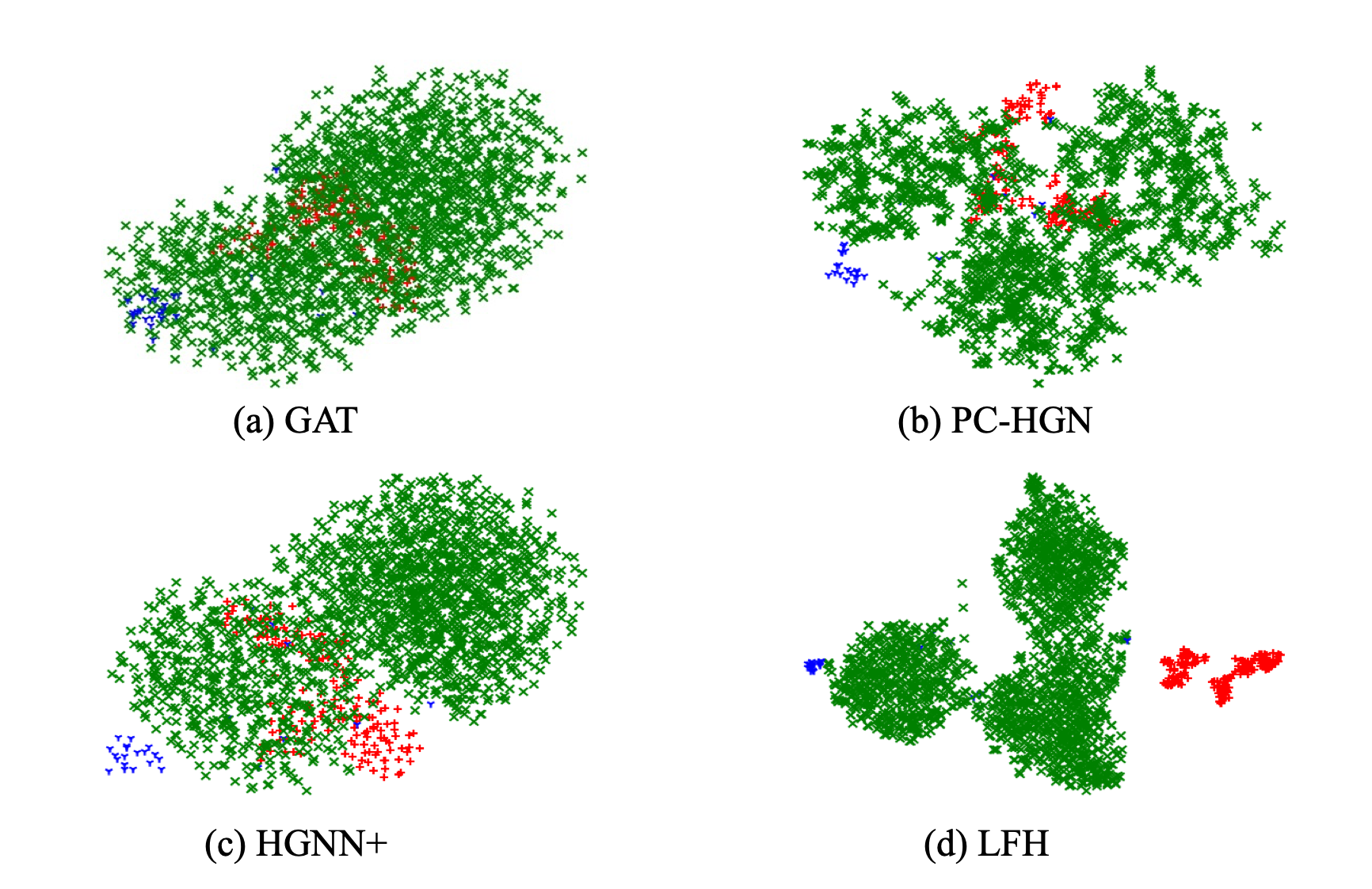}
        	\vspace{-4mm}
		\caption{\rev{A t-SNE visualization of our proposed model compared to baseline models on the DBLP dataset (red indicates paper nodes, green denotes author nodes, and blue represents conference nodes).}}
		\label{fig:vis_dblp}
		\vspace{-5mm}
\end{figure}
\vspace{-2mm}
\subsection{Visualization}~\label{sec:exp.G}
\rev{To investigate the effectiveness of our model compared to other graph learning models in encoding the heterogeneity of the topology, we visualize the learned node embeddings using the t-SNE tool~\cite{van2008visualizing} on the ACM, IMDB and DBLP datasets. It is worth noting that these three datasets naturally possess heterogeneous properties, as represented by various types of nodes (such as \emph{Author}, \emph{Paper} and \emph{Subject} in ACM). Visualizing the distributions of nodes with different types directly demonstrates the method's ability to capture heterogeneity. We compare \pname with other three baselines, GAT, PC-HGN, and HGNN+. These models represent the top-performing baselines for node classification tasks in homogeneous pairwise graph learning, heterogeneous pairwise graph learning, and hypergraph learning. As shown in Fig.~\ref{fig:vis}-\ref{fig:vis_dblp}, our proposed model \pname yields discernible clusters compared to other baseline models across the ACM, IMDB and DBLP datasets, which verifies the effectiveness of the proposed method in capturing the heterogeneity of the topology.}

\begin{figure}[t!]
	\vspace{-5mm}
        \captionsetup{labelfont={color=black}}
        \centering
        \includegraphics[width=0.83\linewidth]{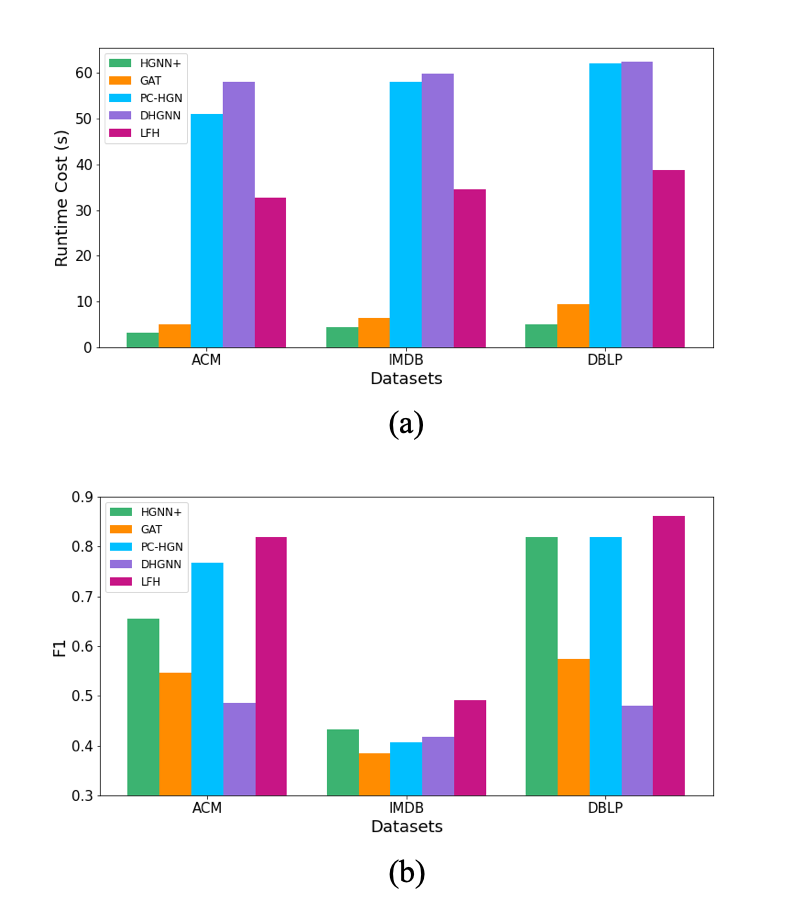}
        	\vspace{-4mm}
		\caption{\rev{(a) Runtime cost with different baseline models on three datasets. (b) Model performance comparison.}}
		\label{fig:time}
		\vspace{-5mm}
\end{figure}
\subsection{Timing Efficiency Analysis}
\rev{We compare the time efficiency on a workstation equipped with an Intel Core i7 CPU @ 2.6 GHz and 32 GB of RAM against four baseline models. DHGNN is a hypergraph model utilizing the cluster method to dynamically construct the hypergraph. The other three baselines, GAT, PC-HGN, and HGNN+, are the top-performing baselines in homogenous pairwise graph learning models, heterogeneous pairwise graph learning models, and hypergraphs learning models in node classification tasks. We set the number of training epochs for all models to 100 to make a fair comparison across three datasets, and we report the results for runtime cost and performance in Fig.~\ref{fig:time}. }

\rev{
From Fig.~\ref{fig:time} (a), it is observed that \pname achieves a significantly lower training time cost compared to DHGNN and PC-HGN on all three datasets. It is worth noting that the runtime reported for our model excludes the time spent on initial node embedding. The encoded pairwise connectivities in pre-trained initial node embedding provide our model with a better starting point from the beginning of the training and thus enhance the learning of representation. As illustrated in Fig.~\ref{fig:time} (b), \pname demonstrates superior performance on all three datasets when compared to all competing baseline models. Apart from that, we observe that both HGNN+ and GAT require less time for training than our model. However, despite time efficiency in the training process, HGNN+ and GAT both fail to capture the heterogeneity, leading to an inferior performance when compared with \pname. The capacity of different models to encode the heterogeneity has been further analyzed in the Section~\ref{sec:exp.G}.
}

\section{conclusion}~\label{sec:conclusion}
In this paper, we propose a heterogenous hypergraph learning framework for representation learning. This framework first generates the high-quality initial node embedding using the designated pairwise fusion function, aiming to exploit the pairwise graph information at most. Afterwards, the multi-type hyperedges are constructed dynamically, forming the hypergraph together. The embedding is then updated iteratively through type-specific attention, aiming to encode the heterogeneous attributes into the embedding space. We conduct comprehensive experiments on three widely-used public datasets and comparison with twelve baseline methods to demonstrate the effectiveness of our proposed framework. The results and analysis reveal that the proposed framework can achieve new state-of-the-art performance on both node classification and link prediction tasks. \rev{In the future, we plan to extend our method to tackle regression problems involving temporal-spatial data by constructing hyperedges that incorporate temporal information, thereby better encoding the evolution of embeddings from this data. Furthermore, we also intend to apply our method to other fields where data naturally have a graph structure, such as protein data in bioinformatics, to provide better representative embeddings for downstream tasks.}

\bibliographystyle{IEEEtran}
\bibliography{IEEEabrv,sample-base}
\vspace{2cm}
\begin{IEEEbiography}[{\includegraphics[width=1in,height=1in,clip,keepaspectratio]{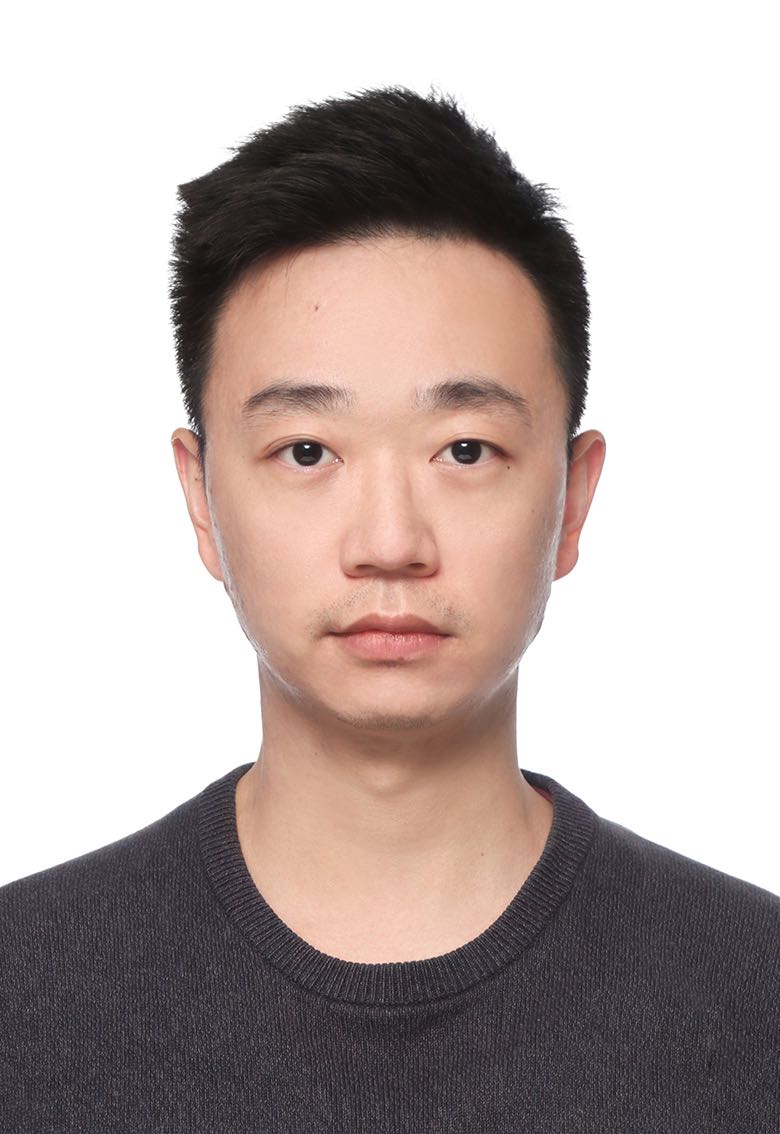}}]{Tiehua Zhang} received the Ph.D. degree in Computer Science from the School of Software and Electrical Engineering, Swinburne University of Technology, Australia, in 2020. He was a Postdoctoral Researcher with the Department of Computing, Macquarie University from 2020 to 2021, a Research Scientist and team leader at Ant Group from 2021 to 2024. He is currently an Assistant Professor with the School of Computer Science and Technology, Tongji University. His research interests encompass collaborative learning/optimization, edge intelligence, graph learning, and the Internet of Things.

\end{IEEEbiography}

\begin{IEEEbiography}[{\includegraphics[width=1in,height=1in,clip,keepaspectratio]{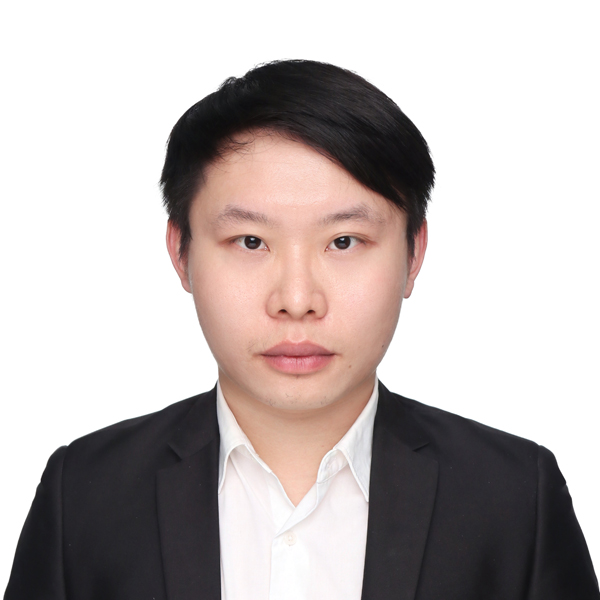}}]{Yuze Liu} received the B.S. degree in Computer Science from Shanghai Normal University, Shanghai, China in 2016. And the M.S. degree in Applied Statistics from Beijing Forestry University, Beijing, China in 2019. He is currently pursuing the Ph.D. degree with the School of Science, Computing and Engineering Technologies, Swinburne University of Technology, Melbourne, Australia. His research interests include federated learning, graph neural networks and natural language processing.

\end{IEEEbiography}

\begin{IEEEbiography}[{\includegraphics[width=1in,height=1in,clip,keepaspectratio]{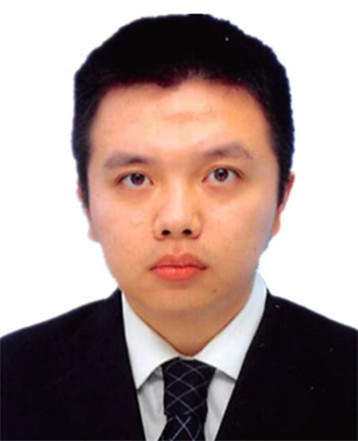}}]{Zhishu Shen} received the B.E. degree from the School of Information Engineering at the Wuhan University of Technology, Wuhan, China, in 2009, and the M.E. and Ph.D. degrees in Electrical and Electronic Engineering and Computer Science from Nagoya University, Japan, in 2012 and 2015, respectively. He is currently an Associate Professor in the School of Computer Science and Artificial Intelligence, Wuhan University of Technology. From 2016 to 2021, he was a research engineer of KDDI Research, Inc., Japan. His major interests include network design and optimization, data learning, edge
computing and the Internet of Things.

\end{IEEEbiography}

\begin{IEEEbiography}[{\includegraphics[width=1in,height=1in,clip,keepaspectratio]{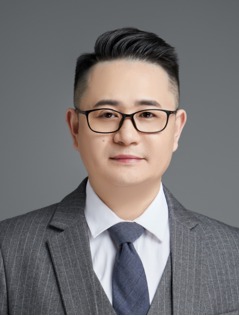}}]{Xingjun Ma} is an Associate Professor in the School of Computer Science at Fudan University. He received his Ph.D. degree from The University of Melbourne. His main research area is trustworthy machine learning, aiming to design secure, robust, explainable, privacy-preserving, and fair machine learning models for diverse AI applications.

\end{IEEEbiography}


\begin{IEEEbiography}[{\includegraphics[width=1in,height=1in,clip,keepaspectratio]{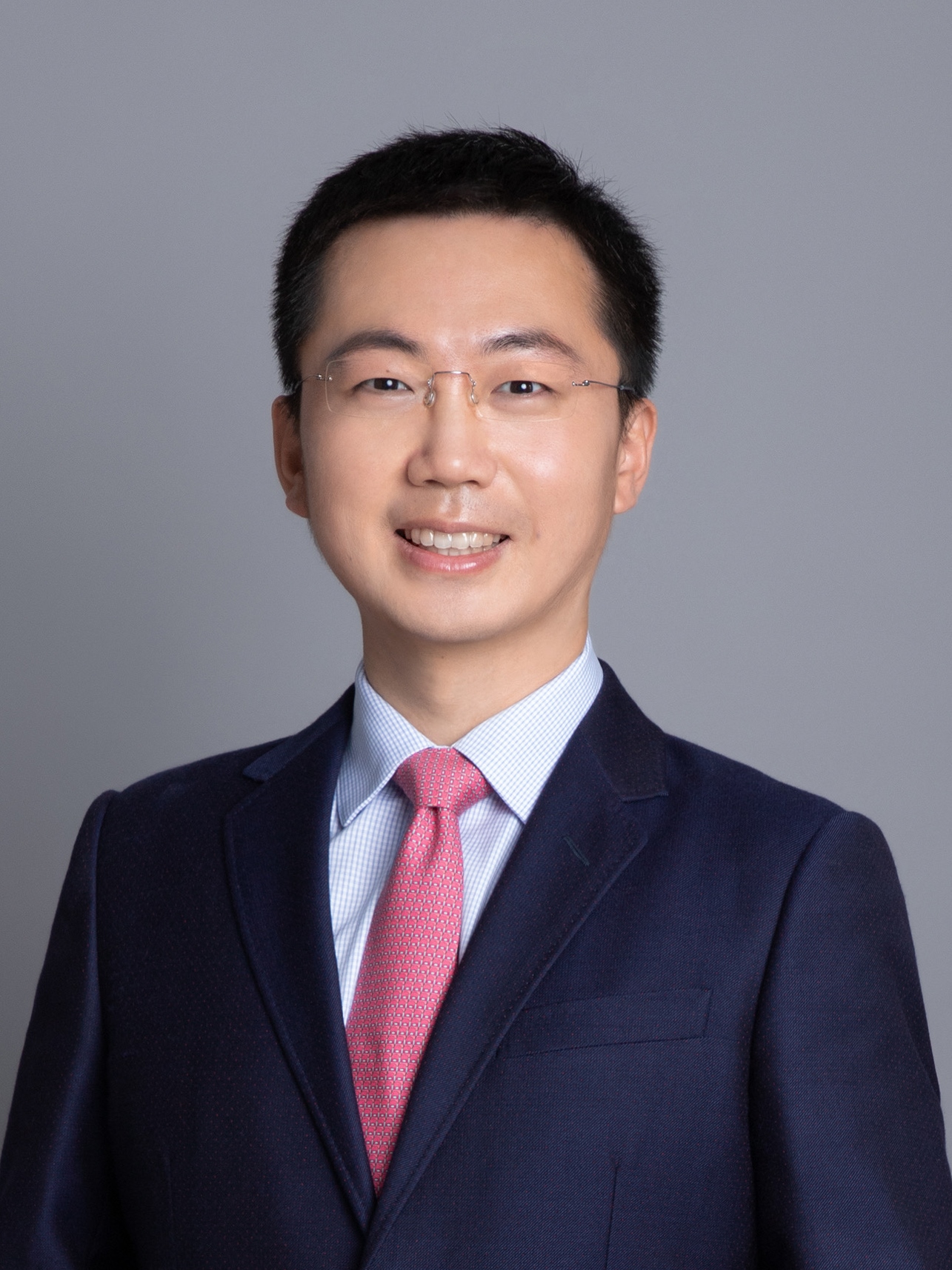}}]{Peng Qi} received the  B.E. degree in automation from Beijing Jiaotong University, Beijing, China, in 2010, the MS degree in electrical engineering from the KTH Royal Institute of Technology, Stockholm, Sweden, in 2012, and the Ph.D. degree in robotics from King’s College London, U.K., in February 2016. He was a research fellow with the National University of Singapore, from September 2015 to August 2016, and a visiting scholar (honored) with The Chinese University of Hong Kong, from September 2016 to February 2017. He is currently an Associate Professor of robotics with Tongji University. His research interests include machine learning, intelligent sensing and interaction, and autonomous intelligent systems.

\end{IEEEbiography}

\begin{IEEEbiography}[{\includegraphics[width=1in,height=1in,clip,keepaspectratio]{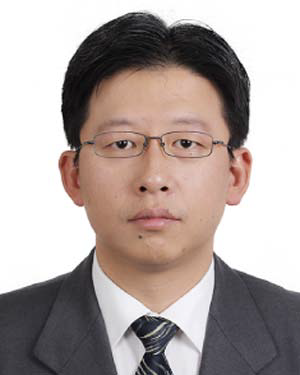}}]{Zhijun Ding} received the Ph.D. degree from Tongji University, Shanghai, China, in 2007. He is currently a Professor with the School of Computer Science and Technology, Tongji University, Shanghai, China. His research interests include
formal method, services computing, and workflow.
He has published more than 100 papers in domestic and international academic journals and
conference proceedings.

\end{IEEEbiography}

\begin{IEEEbiography}[{\includegraphics[width=1in,height=1in,clip,keepaspectratio]{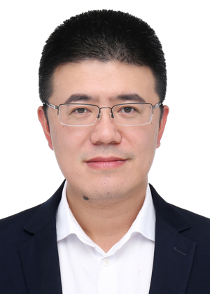}}]{Jiong Jin} received the B.E. degree with First Class Honours in Computer Engineering from Nanyang Technological University, Singapore, in 2006, and the Ph.D. degree in Electrical and Electronic Engineering from the University of Melbourne, Australia, in 2011. He is currently a full Professor in the School of Science, Computing and Engineering Technologies, Swinburne University of Technology, Melbourne, Australia. His research interests include network design and optimization, edge computing and intelligence, robotics and automation, and cyber-physical systems and Internet of Things as well as their applications in smart manufacturing, smart transportation and smart cities. He was recognized as an Honourable Mention in the AI 2000 Most Influential Scholars List in IoT (2021 and 2022). He is currently an Associate Editor of IEEE Transactions on Industrial Informatics. 

\end{IEEEbiography}

\end{document}